\documentclass[10pt,journal,compsoc]{IEEEtran}
\ifCLASSOPTIONcompsoc
  \usepackage[nocompress]{cite}
\else
  \usepackage{cite}
\fi

\usepackage{amsmath,amssymb}
\usepackage{algorithmic}
\usepackage{multirow}
\usepackage[aboveskip=8pt]{caption}
\usepackage[dvips]{graphicx}
\DeclareGraphicsExtensions{.pdf}
\usepackage{tabularx}
\usepackage[bookmarks=false,colorlinks=true,linkcolor=black,citecolor=black,filecolor=black,urlcolor=black]{hyperref}
\usepackage{epsfig}

\usepackage{subfigure}
\usepackage{float}
\usepackage{bbm}
\usepackage{booktabs}
\usepackage{pifont}
\newcommand{\cmark}{\ding{51}}%
\newcommand{\xmark}{\ding{55}}%
\usepackage{threeparttable}
\usepackage{ragged2e}

\begin{document}

\title{Fully Convolutional Networks for Panoptic Segmentation with Point-based Supervision}

\author{Yanwei~Li,
        Hengshuang~Zhao,
        Xiaojuan~Qi,
        Yukang~Chen,
        Lu~Qi, \\
        Liwei~Wang, 
        Zeming~Li,
        Jian~Sun,
        Jiaya~Jia,~\IEEEmembership{Fellow,~IEEE}
\IEEEcompsocitemizethanks{
\IEEEcompsocthanksitem Y.~Li, Y.~Chen, L.~Qi, L.~Wang and J.~Jia are with the Department of Computer Science and Engineering, The Chinese University of Hong Kong. Email: \{ywli, yukangchen, luqi, lwwang, leojia\}@cse.cuhk.edu.hk
\IEEEcompsocthanksitem H.~Zhao is with the Department of Engineering Science, University of Oxford. Email: hengshuang.zhao@eng.ox.ac.uk
\IEEEcompsocthanksitem X.~Qi is with the Department of Electrical and Electronic Engineering, The University of Hong Kong. Email: xjqi@eee.hku.hk
\IEEEcompsocthanksitem Z.~Li and J.~Sun are with the MEGVII Technology. Email: \{lizeming, sunjian\}@megvii.com
}
}


\IEEEtitleabstractindextext{%
    \begin{abstract} \justifying
    In this paper, we present a conceptually simple, strong, and efficient framework for fully- and weakly-supervised panoptic segmentation, called Panoptic FCN. 
    Our approach aims to represent and predict foreground things and background stuff in a unified fully convolutional pipeline, which can be optimized with point-based fully or weak supervision. 
    In particular, Panoptic FCN encodes each object instance or stuff category with the proposed kernel generator and produces the prediction by convolving the high-resolution feature directly. 
    With this approach, instance-aware and semantically consistent properties for things and stuff can be respectively satisfied in a simple generate-kernel-then-segment workflow. 
    Without extra boxes for localization or instance separation, the proposed approach outperforms the previous box-based and -free models with high efficiency. 
    Furthermore, we propose a new form of point-based annotation for weakly-supervised panoptic segmentation. It only needs several random points for both things and stuff, which dramatically reduces the annotation cost of human.
    The proposed Panoptic FCN is also proved to have much superior performance in this weakly-supervised setting, which achieves 82\% of the fully-supervised performance with only 20 randomly annotated points per instance.
    Extensive experiments demonstrate the effectiveness and efficiency of Panoptic FCN on COCO, VOC 2012, Cityscapes, and Mapillary Vistas datasets. 
    And it sets up a new leading benchmark for both fully- and weakly-supervised panoptic segmentation.
    Our code and models are made publicly available at \href{https://github.com/dvlab-research/PanopticFCN}{https://github.com/dvlab-research/PanopticFCN}.
    \end{abstract}
    
}

\maketitle

\IEEEdisplaynontitleabstractindextext
\IEEEpeerreviewmaketitle

\begin{figure}[t!] 

\centering
\subfigure[Separate representation]{
\includegraphics[width=0.45\linewidth]{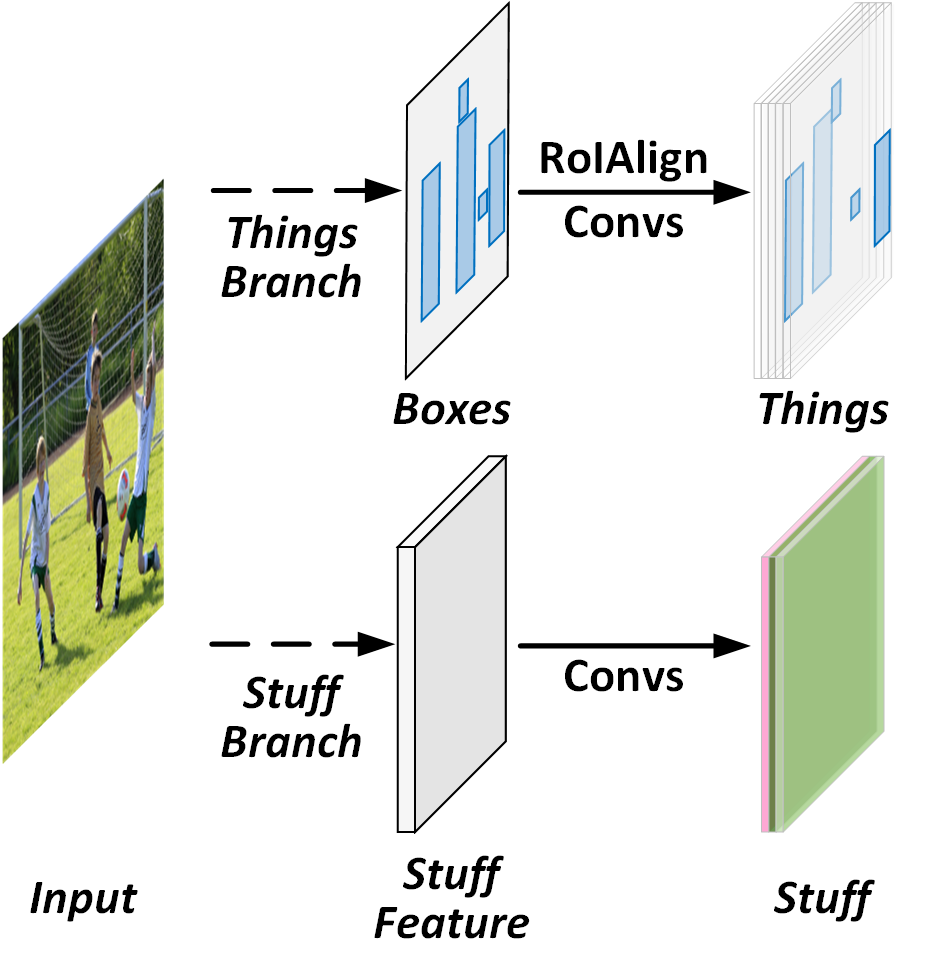}
\label{fig:intro_trad}
}
\subfigure[Unified representation]{
\includegraphics[width=0.45\linewidth]{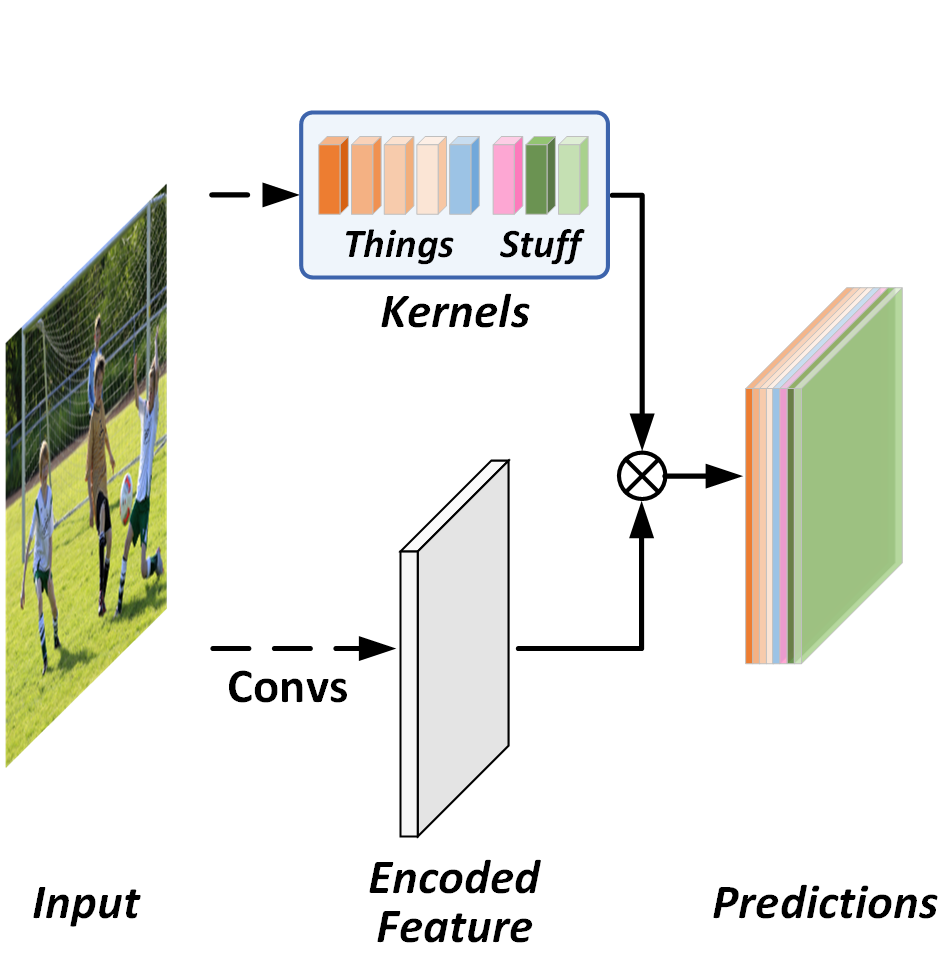}
\label{fig:intro_ours}
}
\caption{Compared with previous method, which often utilizes separate branches to handle {\em things} and {\em stuff} in~\ref{fig:intro_trad}, Panoptic FCN~\ref{fig:intro_ours} represents {\em things} and {\em stuff} uniformly with generated kernels. Here, an example with box-based stream for {\em things} and FCN-based branch for {\em stuff} is given in~\ref{fig:intro_trad}. The shared backbone is omitted for concision.}
\vspace{1em}
\label{fig:intro}
\end{figure}

\IEEEraisesectionheading{\section{Introduction}}
\IEEEPARstart{P}{anoptic} segmentation, aiming to assign each pixel with a semantic label and unique identity, is regarded as a challenging task. In panoptic segmentation~\cite{kirillov2019panoptic}, countable and uncountable instances ({\em i.e.,} things and stuff) are expected to be represented and resolved in a unified workflow. One main difficulty impeding unified representation for fully- and weakly-supervised panoptic segmentation comes from conflicting properties requested by things and stuff. Specifically, to distinguish among various identities, countable things usually rely on {\em instance-aware} features, which vary with objects. In contrast, uncountable stuff prefers {\em semantically consistent} characters, which ensures consistent predictions for pixels with the same semantic meaning.
An example is given in Fig.~\ref{fig:intro}, where embedding of {\em individuals} should be diverse for inter-class variations, while characters of {\em grass} should be similar for intra-class consistency.

For conflict at the feature level, specific modules are usually tailored for things and stuff separately, as presented in Fig.~\ref{fig:intro_trad}. In particular, {\em instance-aware} demand of things is satisfied mainly from two streams, namely box-based~\cite{kirillov2019panopticfpn, xiong2019upsnet, li2019attention} and box-free~\cite{yang2019deeperlab, gao2019ssap, cheng2020panoptic} methods. Meanwhile, the {\em semantic-consistency} of stuff is met in a pixel-by-pixel manner~\cite{long2015fully}, where similar semantic features would bring identical predictions. A classic case is Panoptic FPN~\cite{kirillov2019panopticfpn}, which utilizes Mask R-CNN~\cite{he2017mask} and FCN~\cite{long2015fully} in separated branches to respectively classify things and stuff, similar to that of Fig.~\ref{fig:intro_trad}. Although attempt~\cite{yang2019deeperlab, gao2019ssap, cheng2020panoptic} has been made to predict things without boxes, extra predictions ({\em e.g.,} affinities~\cite{gao2019ssap}, and offsets~\cite{yang2019deeperlab}) together with post-process procedures are still needed to distinguish among instances. 
This slows down the whole system and hinders it from being fully convolutional. 
Therefore, box-level annotation~\cite{dai2015boxsup,song2019box,tian2021boxinst} and point-based manners~\cite{maninis2018deep,papadopoulos2017extreme,bearman2016s} are utilized to provide weak supervision for things and stuff. Consequently, a unified representation is required to bridge this gap for both fully- and weakly-supervised settings. 

In this paper, we propose a fully convolutional framework for unified representation, called {\em Panoptic FCN}. In particular, Panoptic FCN encodes each instance into a specific kernel and generates the prediction by convolutions directly. Thus, both things and stuff can be predicted together with the same resolution. In this way, {\em instance-aware} and {\em semantically consistent} properties for things and stuff can be respectively satisfied in a unified workflow, which is briefly illustrated in Fig.~\ref{fig:intro_ours}. To sum up, the key idea of Panoptic FCN is to {\em represent and predict things and stuff uniformly with generated kernels in a fully convolutional pipeline}. 

\begin{figure*}[!t]
\centering
\includegraphics[width=0.97\linewidth]{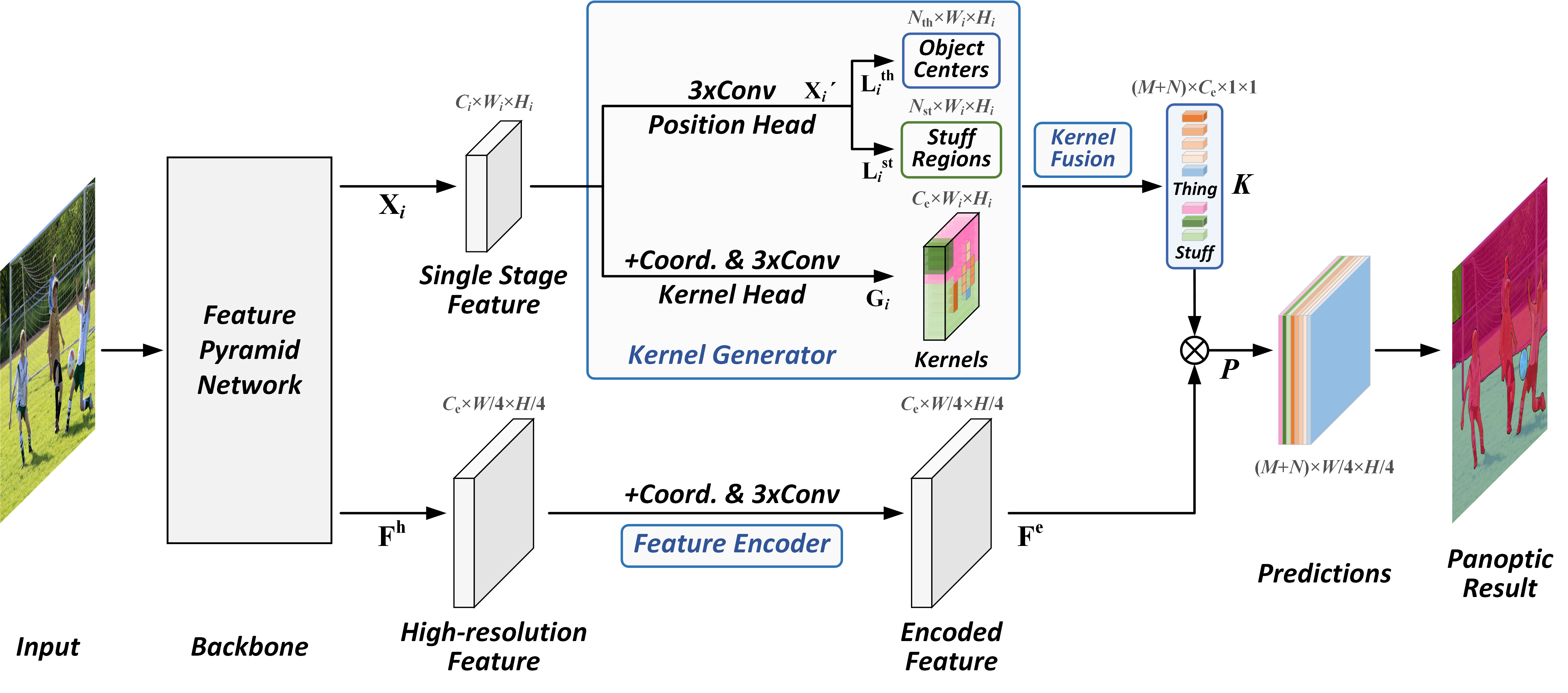} 
\caption{{\bf The framework of Panoptic FCN}. The proposed framework mainly contains {\em three} components, namely {\em kernel generator}, {\em kernel fusion}, and {\em feature encoder}. In {\em kernel generator}, position head is designed to locate and classify object centers along with stuff regions; kernel head in each stage is used to generate kernel weights for both things and stuff. Then, {\em kernel fusion} is utilized to merge kernel weights with the same identity from different stages. And {\em feature encoder} is adopted to encode the high-resolution feature with details. With the generated kernel weight for each instance, both things and stuff can be predicted with a simple convolution directly.}
\label{fig:main}
\end{figure*}

To this end, {\em kernel generator} and {\em feature encoder} are respectively designed for {\em kernel weights generation} and {\em shared feature encoding}. 
In kernel generator, we draw inspirations from point-based object detectors~\cite{law2018cornernet, zhou2019objects} and utilize the position head to locate and classify foreground objects, as well as background stuff, by {\em object centers} and {\em stuff regions}, as demonstrated in Fig.~\ref{fig:main}. 
Then, we select kernel weights~\cite{jia2016dynamic} with the same positions from the kernel head to represent instances. 
For the {\em instance-awareness} and {\em semantic-consistency}, a kernel-level operation, called {\em kernel fusion}, is further proposed, which merges kernel weights that are predicted to have the same identity or semantic category. 
With a naive feature encoder that preserves the high-resolution feature with details, each prediction of things and stuff is produced by convolving with generated kernels directly.

In general, the proposed method is distinguished from two aspects. Firstly, different from previous work for {\em things} generation~\cite{he2017mask,chen2019tensormask,wang2019solo}, which outputs dense predictions and then utilizes NMS for overlaps removal, the deigned framework generates {\em instance-aware} kernels and produces each specific instance directly. Moreover, compared with traditional FCN-based methods for {\em stuff} prediction~\cite{long2015fully,zhao2017pyramid,chen2018encoder}, which select the most likely category in a pixel-by-pixel manner, our approach aggregates global context into {\em semantically consistent} kernels and presents results of existing semantic classes in a whole-instance manner.

Thanks to the designed workflow, we can easily extend it to support the weakly-supervised setting and save tremendous annotation costs. Instead of using box-based annotation strategies~\cite{li2018weakly,tian2021boxinst,cheng2021pointly} that are tailored for objects, we propose a unified manner for both things and stuff annotation, namely {\em random points}. In particular, we randomly annotate several points inside each instance for things and stuff, no matter locate in the region boundary~\cite{papadopoulos2017extreme} or not.
With the randomly annotated points, we further investigate the strategies for target generation and shape augmentation in Sec.~\ref{sec:point_supervision}. 
Generally, the proposed pointly-supervised method reduces the annotation cost to 27\% while maintains 82\% of the fully-supervised performance on COCO dataset.

The overall approach can be easily instantiated for fully- and weakly-supervised panoptic segmentation. 
To demonstrate its superiority, we give extensive ablation studies in Sec.~\ref{sec:ablation} and Sec.~\ref{sec:ablation_point}. 
Furthermore, experimental results are reported on COCO~\cite{kirillov2019panoptic,lin2014microsoft,caesar2018coco}, VOC 2012~\cite{everingham2015pascal}, Cityscapes~\cite{cordts2016cityscapes}, and Mapillary Vistas~\cite{neuhold2017mapillary} datasets. 
Without bells-and-whistles, Panoptic FCN outperforms previous methods with efficiency, and attains {\bf 52.1}\% PQ and {\bf 52.7}\% PQ on COCO {\em val} and {\em test-dev} set. 
Meanwhile, it surpasses all similar {\em box-free} methods by large margins and achieves cutting-edge results on Cityscapes and Mapillary Vistas {\em val} set with {\bf 65.9}\% PQ and {\bf 45.7}\% PQ, respectively. 
In the weakly-supervised setting, the proposed manner achieves leading performance on COCO and VOC 2012 datasets, and maintains over 80\% performance of the fully-supervised version with only 20 random points for annotation.

\vspace{0.5em}
\noindent
\textbf{Difference from our conference paper.}
This manuscript significantly improves the conference version~\cite{li2021panopticfcn}: 
(\romannumeral1) we introduce a pure point-based format for weakly-supervised panoptic segmentation, reducing the human annotation cost greatly; (\romannumeral2) we investigate the annotation and generation strategies for point-based supervision with effectiveness and efficiency; (\romannumeral3) we develop a morphological augmentation for point-based annotation and achieve significant improvement; (\romannumeral4) we conduct additional experiments and analysis for both fully- and pointly-supervised Panoptic FCN on VOC 2012~\cite{everingham2015pascal} dataset; (\romannumeral5) we empirically show that Panoptic FCN can be incorporated with recently proposed Swin Transformer~\cite{liu2021Swin} to further improve the results and set up new benchmarks on several widely-adopted datasets.

\section{Related Work}~\label{sec:related_work}
In this section, we first review recent developments in relevant segmentation tasks. Then, we introduce advances in weakly- and pointly-supervised methods.

\vspace{0.5em}
\noindent
\textbf{Panoptic segmentation.}
Traditional approaches mainly conduct segmentation for things and stuff separately. The benchmark for panoptic segmentation~\cite{kirillov2019panoptic,lin2014microsoft,caesar2018coco} directly combines predictions of things and stuff from different models, causing heavy computational overhead. To solve this problem, methods have been proposed by dealing with things and stuff in one model but in separate branches, including Panoptic FPN~\cite{kirillov2019panopticfpn}, AUNet~\cite{li2019attention}, and UPSNet~\cite{xiong2019upsnet}. 
From the view of instance representation, previous work mainly formats things and stuff from different perspectives. Foreground things are usually separated and represented with boxes~\cite{kirillov2019panopticfpn,yang2019sognet,chen2020banet,li2020unifying} or aggregated according to center offsets~\cite{yang2019deeperlab}, while background stuff is often predicted with a parallel FCN~\cite{long2015fully} branch. Although methods of~\cite{li2018weakly,gao2019ssap} represent things and stuff uniformly, the inherent ambiguity cannot be resolved well merely with the pixel-level affinity, which yields the performance drop in complex scenarios. In contrast, the proposed Panopic FCN represents things and stuff in a uniform and fully convolutional framework with decent performance and efficiency.

\vspace{0.5em}
\noindent
\textbf{Instance segmentation.}
Instance segmentation aims to discriminate objects in the pixel level, which is a finer representation compared with detected boxes. For {\em instance-awareness}, previous works can be roughly divided into two streams, {\em i.e.,} box-based methods and box-free approaches. Box-based methods usually utilize detected boxes to locate or separate objects~\cite{he2017mask,liu2018path,bolya2019yolact,lee2020centermask,qi2020pointins}. A classic example in this stream is Mask R-CNN~\cite{he2017mask}, which extends Faster R-CNN~\cite{ren2015faster} with an extra mask branch and conduct object segmentation within the detected boxes using this branch. Meanwhile, box-free approaches are designed to generate instances without the assistance of object boxes~\cite{gao2019ssap,chen2019tensormask,wang2019solo,wang2020solov2}. Recently, AdaptIS~\cite{sofiiuk2019adaptis} and CondInst~\cite{tian2020conditional} are proposed to utilize point-proposal for instance segmentation. However, instance aggregation or object-level removal is still needed for results. In this paper, we represent objects in a box-free pipeline, which generates the kernel and produces results by convolving the detail-rich feature directly, with no need for object-level removal~\cite{hu2018relation,qi2018sequential}.

\vspace{0.5em}
\noindent
\textbf{Semantic segmentation.}
Semantic segmentation assigns each pixel with a semantic category, without considering diverse object identities. In recent years, rapid progress has been made on top of FCN~\cite{long2015fully}. Due to the {\em semantically consistent} property, several attempts have been made to capture contextual cues from wider perception fields~\cite{zhao2017pyramid,chen2017rethinking,chen2018encoder} or establish pixel-wise relationship for long-range dependencies~\cite{zhao2018psanet,huang2019ccnet,song2019learnable}. More recently, visual attention mechanism~\cite{fu2019dual,yuan2020object} and vision transformer~\cite{dosovitskiy2020image,zheng2021rethinking} are introduced to enhance relation modeling. However, these works still view semantic segmentation as per-pixel classification, regardless of mask prediction for each category. There is also work to design network architectures for semantic segmentation automatically~\cite{liu2019auto,li2020learning}, which is beyond the scope of this paper. Our proposed Panoptic FCN adopts a similar method to represent things and stuff, which aggregates global context into a specific kernel to predict corresponding semantic category.

\vspace{0.5em}
\noindent
\textbf{Weakly-supervised segmentation.}
Previous work for weakly-supervised approaches utilize image-level labels, bounding boxes, or other readily accessible manners to provide annotation for segmentation tasks. With image-level labels, they usually use segment proposals~\cite{pont2016multiscale,shen2021toward} or generated pseudo-label~\cite{li2018weakly,ahn2019weakly,arun2020weakly} to provide supplemental target for network optimization. However, the image-level labels, as well as generated pseudo label, can not always provide high-quality annotations, especially in complex scenes like COCO~\cite{kirillov2019panoptic,lin2014microsoft,caesar2018coco}.
Bounding box is another widely-used supervision because of the easy accessibility. Compared with image-level labels only, bounding boxes can provide not only category information but also the location and tight bound. There are several approaches~\cite{dai2015boxsup,khoreva2017simple,song2019box,hsu2019weakly,kulharia2020box2seg} for box-supervised segmentation and achieve satisfactory results. More recently, BoxInst~\cite{tian2021boxinst} is proposed to utilize projection loss and pixel similarity based on CondInst~\cite{tian2020conditional}. However, the box-based supervision requires time-consuming extreme points that are tailored for objects and can not suit the complex stuff with arbitrary shape in panoptic segmentation. 
There are also some works using other annotation manners, like scribble~\cite{lin2016scribblesup} and  points~\cite{bearman2016s,maninis2018deep,papadopoulos2017extreme}, which are discussed in the following part.

\vspace{0.5em}
\noindent
\textbf{Point-based supervision.}
Compared with image-level and box-based annotations, points provide more accurate location and positive region within object without bringing much cost. Traditional approaches mainly utilize extreme points~\cite{maninis2018deep,papadopoulos2017extreme} and individual points~\cite{bearman2016s} for optimization. However, the extreme points actually consume much more time than individual points~\cite{cheng2021pointly} during annotation. And the single point for each instance~\cite{bearman2016s} can not provide relative precise target for practical usage.
Most recently,~\cite{cheng2021pointly} is designed to provide positive and negative points within each object boxes for instance segmentation. However, as declared before, such approach requires pre-annotated boxes and can not suit stuff region in panoptic segmentation. Different from above mentioned approaches, the proposed method only utilizes random points for target generation and network optimization of things and stuff.

\section{Panoptic FCN} \label{sec:methods}
Panoptic FCN is conceptually simple: {\em kernel generator} is introduced to generate kernel weights for things and stuff with different categories; {\em kernel fusion} is designed to merge kernel weights with the same identity from multiple stages; and {\em feature encoder} is utilized to encode the high-resolution feature. In this section, we first elaborate the above components. Then, we further introduce the strategy to weakly-supervised panoptic segmentation with {\em point-based} supervision. The network details are given at the end of this section. 

\subsection{Kernel Generator}
To generate a specific kernel for each instance of thing and stuff, we utilize {\em position head} to locate the center of each object and the region of each stuff category.
Meanwhile, a parallel {\em kernel head} is designed to provide candidate kernel features, which are further selected and fused for instance specificity in the following kernel fusion.
Given the feature $\mathbf{X}_i$ from the $i$-th stage in FPN~\cite{lin2017feature}, the proposed kernel generator aims at generating the kernel weight map $\mathbf{G}_i$ with positions for things $\mathbf{L}_i^{\mathrm{th}}$ and stuff $\mathbf{L}_i^{\mathrm{st}}$, as depicted in Fig.~\ref{fig:main}. 

\subsubsection{Position Head.} ~\label{sec:position_head}
With the input $\mathbf{X}_{i}\in \mathbb{R}^{C_i\times W_i\times H_i}$, we adopt stacks of convolutions to encode the feature map and generate $\mathbf{X}'_{i}$, as presented in Fig.~\ref{fig:main}. 
Then we need to locate and classify each instance from the shared feature map $\mathbf{X}'_{i}$. 
However, according to the definition~\cite{heitz2008learning,kirillov2019panoptic}, things can be distinguished by object centers, while stuff is uncountable. 
Thus, we adopt {\em object centers} and {\em stuff regions} to respectively represent position of each individual and stuff category. 
It means regions with the same semantic meaning are viewed as one instance.
In particular, object map $\mathbf{L}_i^{\mathrm{th}}\in \mathbb{R}^{N_{\mathrm{th}}\times W_i\times H_i}$ and stuff map $\mathbf{L}_i^{\mathrm{st}}\in \mathbb{R}^{N_{\mathrm{st}} \times W_i\times H_i}$ are generated by convolutions directly with the shared feature map $\mathbf{X}'_{i}$, where $N_{\mathrm{th}}$ and $N_{\mathrm{st}}$ denote the number of semantic category for things and stuff. 

To better optimize $\mathbf{L}_i^{\mathrm{th}}$ and $\mathbf{L}_i^{\mathrm{st}}$, different strategies are adopted to generate the ground-truth. For the $k$-th object in class $c$, we split positive key-points onto the $c$-th channel of the heatmap $\mathbf{Y}_i^{\mathrm{th}}\in\left[0,1\right]^{N_{\mathrm{th}}\times W_i\times H_i}$ with Gaussian kernel, similar to that in~\cite{law2018cornernet, zhou2019objects}. 
Thus, ground-truth of the $k$-th object in class $c$ can be represented as
\begin{equation}\label{equ:thing_gt}
\mathbf{Y}_{i,c,x,y}^{\mathrm{th}}=\exp\left(-\frac{{(x-{\widetilde x}_k)}^2+{(y-{\widetilde y}_k)}^2}{2\sigma_k^2}\right),
\end{equation}
where $({\widetilde x}_k, {\widetilde y}_k)$ denotes center of the $k$-th object, and $\sigma_k$ indicates the object size-adaptive standard deviation in~\cite{law2018cornernet}. 
In this work, $\sigma_k$  is set to $(2r+1)/3$, where $r$ denotes the gaussian radius of the $k$-th object.
With respect to stuff, we produce the ground-truth~$\mathbf{Y}_i^{\mathrm{st}}\in\left[0,1\right]^{N_{\mathrm{st}}\times W_i\times H_i}$ by bilinear interpolating the one-hot semantic label to corresponding sizes. Hence, the position head can be optimized with

\begin{equation} ~\label{equ:loc}
\begin{aligned}
{\mathcal L}_{\mathrm{pos}}^{\mathrm{th}}=&\sum_i\mathrm{FL}(\mathbf{L}_i^{\mathrm{th}},\mathbf{Y}_i^{\mathrm{th}})/N_{\mathrm{th}}, \\
{\mathcal L}_{\mathrm{pos}}^{\mathrm{st}}=&\sum_i\mathrm{FL}(\mathbf{L}_i^{\mathrm{st}},\mathbf{Y}_i^{\mathrm{st}})/({W_i\times H_i}), \\
\end{aligned}
\end{equation}
where ${\mathcal L}_{\mathrm{pos}}^{\mathrm{th}}$ and ${\mathcal L}_{\mathrm{pos}}^{\mathrm{st}}$ denote the loss function for object centers and stuff regions, respectively. 
And $\mathrm{FL}(\cdot,\cdot)$ represents the Focal Loss~\cite{lin2017focal} for corresponding prediction $\mathbf{L}_i$ whose ground-truth is $\mathbf{Y}_i$ in the $i$-th stage. 
Consequently, the loss function for position head is formulated as
\begin{equation} ~\label{equ:pos}
{\mathcal L}_{\mathrm{pos}} = {\mathcal L}_{\mathrm{pos}}^{\mathrm{th}} + {\mathcal L}_{\mathrm{pos}}^{\mathrm{st}}.
\end{equation}
For inference, $D^{\mathrm{th}}_i=\left\{(x,y):\mathbbm{1}(\mathbf{L}^{\mathrm{th}}_{i,c,x,y})=1\right\}$ and $D^{\mathrm{st}}_i=\left\{(x,y):\mathbbm{1}(\mathbf{L}^{\mathrm{st}}_{i,c,x,y})=1\right\}$ are selected to represent the existence of object centers and stuff regions in corresponding positions.
In particular, the indicator for things $\mathbbm{1}(\mathbf{L}^{\mathrm{th}}_{i,c,x,y})$ is marked as positive if point $(x,y)$ in the $c$-th channel is preserved as the peak point. 
Similarly, the indicator for stuff regions $\mathbbm{1}(\mathbf{L}^{\mathrm{st}}_{i,c,x,y})$ is viewed as positive if point $(x,y)$ with category $c$ is selected with the top score among all classes. 
Here, we use $O_i$ to represent the set of all the predicted category $c$.
This process is further explained in Sec.~\ref{sec:train_infer}.

\subsubsection{Kernel Head}
In kernel head, we first capture spatial cues by concatenating relative coordinates to the feature $\mathbf{X}_{i}$, which is similar with CoordConv~\cite{liu2018intriguing}. With the concatenated feature map $\mathbf{X}''_{i}\in \mathbb{R}^{(C_i+2)\times W_i\times H_i}$, stacks of convolutions are adopted to generate the kernel weight map $\mathbf{G}_{i}\in \mathbb{R}^{C_\mathrm{e}\times W_i\times H_i}$, as presented in Fig.~\ref{fig:main}. 
Given predictions $D^{\mathrm{th}}_i$ and $D^{\mathrm{st}}_i$ from the position head, kernel weights with the same coordinates in $\mathbf{G}_{i}$ are chosen to represent corresponding instances. 
For example, assuming candidate $(x_{c}, y_{c})\in D^{\mathrm{th}}_{i}$, kernel weight $\mathbf{G}_{i, :, x_{c}, y_{c}}\in \mathbb{R}^{C_\mathrm{e}\times1\times1}$ is selected to create representation for each object instance.
The same is true for $D^{\mathrm{st}}_i$. 
We denote the selected kernel weights in $i$-th stage for things and stuff as $G^{\mathrm{th}}_i$ and $G^{\mathrm{st}}_i$, respectively. 
In this manner, the kernel weight $G^{\mathrm{th}}_i$ and $G^{\mathrm{st}}_i$ are used to provide kernel-level representation in the $i$-th stage, which are further utilized to produce the instance for each thing and stuff in the following part.

\subsection{Kernel Fusion} ~\label{sec:kernel_fusion}
The kernel fusion is proposed to merge repetitive kernel weights in kernel generator before final instance generation, which guarantees {\em instance-awareness} and {\em semantic-consistency} for things and stuff, respectively. 
In particular, given aggregated kernel weights $G^{\mathrm{th}}$ and $G^{\mathrm{st}}$ from all the stages, the $j$-th kernel weight $K_{j}\in \mathbb{R}^{C_\mathrm{e}\times1\times1}$ is achieved by
\begin{equation}
K_{j}=\mathrm{AvgCluster}(S_{j}),
\end{equation}
where $\mathrm{AvgCluster}$ denotes a simple average-clustering operation that returns an average kernel weight $K_{j}$.
And the candidate set $S_{j}=\{G_{m}:\mathrm{ID}(G_{m})=\mathrm{ID}(G_{j})\}$ includes all the kernel weights, which are predicted to have the same identity $\mathrm{ID}$ with $G_{j}$. 
Here, $G_{m}\in \mathbb{R}^{C_\mathrm{e}\times1\times1}$ denotes the $m$-th candidate in the kernel set $G$.
For object centers, kernel weight $G^{\mathrm{th}}_{m}$ is viewed as identical with $G^{\mathrm{th}}_{j}$ if the {\em cosine similarity} between them surpasses a given threshold $thres$, which will be further investigated in Table~\ref{tab:abla_similar}. For stuff regions, all kernel weights in $G^{\mathrm{st}}$, which share a same category with $G^{\mathrm{st}}_{j}$, are marked as one identity $\mathrm{ID}$. 

With the proposed approach, each kernel weight $K^{\mathrm{th}}_j$ in $K^{\mathrm{th}}=\{K^{\mathrm{th}}_1, ..., K^{\mathrm{th}}_m\}\in \mathbb{R}^{M\times C_\mathrm{e}\times 1 \times 1}$ can be viewed as an embedding for single object, where the total number of objects is $M$. Therefore, kernels with the same identity are merged as a single embedding for things, and each kernel in $K^{\mathrm{th}}$ represents an individual object, which satisfies the {\em instance-awareness} for things. Meanwhile, kernel weight $K^{\mathrm{st}}_j$ in $K^{\mathrm{st}}=\{K^{\mathrm{st}}_1, ..., K^{\mathrm{st}}_n\} \in \mathbb{R}^{N\times C_\mathrm{e}\times 1 \times 1}$ represents the embedding for all $j$-th class pixels, where the existing number of stuff is $N$. With this method, kernels with the same semantic category are fused into a single embedding, which guarantees the {\em semantic-consistency} for stuff. 

\subsection{Feature Encoder} ~\label{sec:feature_encoder}
To preserve details for each instance, high-resolution feature $\mathbf{F}^{\mathrm{h}}\in\mathbb{R}^{C_\mathrm{e}\times W/4\times H/4}$ is utilized for encoding. Feature $\mathbf{F}^{\mathrm{h}}$ can be generated from FPN in several ways, {\em e.g.,} P2 stage feature, summed features from all stages, and features from semantic FPN~\cite{kirillov2019panopticfpn}. These methods are compared in Table~\ref{tab:abla_encoder}. Given the feature $\mathbf{F}^{\mathrm{h}}$, a similar strategy with that in kernel head is applied to encode positional cues and generate the encoded feature $\mathbf{F}^{\mathrm{e}}\in\mathbb{R}^{C_\mathrm{e}\times W/4\times H/4}$, as depicted in Fig.~\ref{fig:main}. Thus, given $M$ and $N$ kernel weights for things $K^{\mathrm{th}}$ and stuff $K^{\mathrm{st}}$ from the kernel fusion, each instance is produced by $\mathbf{P}_j=K_j\otimes \mathbf{F}^{\mathrm{e}}$.
Here, $\mathbf{P}_j$ denotes the $j$-th prediction, and $\otimes$ indicates the convolution. 
That means $M+N$ kernel weights generate $M+N$ instance predictions with resolution $W/4\times H/4$ for the whole image. Consequently, the panoptic result can be produced with a simple process~\cite{kirillov2019panopticfpn}.

\subsection{Training and Inference} \label{sec:train_infer}
In this section, we first introduce the designed weighted dice loss and the final objective function. Then, the detailed scheme for inference is given for clear illustration.
\subsubsection{Training Scheme}
In the training stage, the central point in each object and all the points in stuff regions are utilized to generate kernel weights for things and stuff, respectively. Here, Dice Loss~\cite{milletari2016v} is adopted to optimize the predicted segmentation
\begin{equation} ~\label{equ:dice_raw}
{\mathcal L}_{\mathrm{seg}}=\sum_j\mathrm{Dice}(\mathbf{P}_j,\mathbf{Y}_j^{\mathrm{seg}})/(M+N), \\
\end{equation}
where $\mathbf{Y}_j^{\mathrm{seg}}$ denotes ground-truth for the $j$-th prediction $\mathbf{P}_j$. To further release the potential of kernel generator, multiple positives inside each object are sampled for instance representation. 
In particular, we select $k$ positions with top predicted scores $s$ inside each object in $\mathbf{L}_i^{\mathrm{th}}$, resulting in $k\times M$ kernels, as well as instances. Thus, we generate the same ground-truth $\mathbf{Y}_j^\mathrm{seg}$ for $k$ results in $\mathbf{P}_j$. This will be explored experimentally in Table~\ref{tab:abla_dice}.
For stuff regions, we merge all positions with the same category in each stage using $\mathrm{AvgCluster}$, resulting in a specific kernel. Hence, the factor $k$ can be viewed as 1 for each stuff region, which means all the points in same category are equally treated. 
Then, we replace the original loss with a weighted version
\begin{equation}
\mathrm{WDice}(\mathbf{P}_j,\mathbf{Y}_j^{\mathrm{seg}})=\sum_{k}w_k\mathrm{Dice}(\mathbf{P}_{j,k},\mathbf{Y}_{j}^{\mathrm{seg}}), \\
\end{equation}
where $w_k$ denotes the $k$-th weighted score with $w_k=s_k/\sum_{i}s_i$. According to Eqs.~\eqref{equ:loc} and~\eqref{equ:dice_raw}, optimized target ${\mathcal L}$ is defined with the weighted Dice Loss ${\mathcal L}_{\mathrm{seg}}$ as
\begin{equation} ~\label{equ:dice_weighted}
{\mathcal L}_{\mathrm{seg}}=\sum_j\mathrm{WDice}(\mathbf{P}_j,\mathbf{Y}_j^{\mathrm{seg}})/(M+N), \\
\end{equation}
\begin{equation}~\label{equ:objective_loss}
{\mathcal L}=\lambda_{\mathrm{pos}}{\mathcal L}_{\mathrm{pos}} + \lambda_{\mathrm{seg}}{\mathcal L}_{\mathrm{seg}}.
\end{equation}

\subsubsection{Inference Scheme} 
During inference, Panoptic FCN follows a simple {\em generate-kernel-then-segment} pipeline.
Specifically, we first aggregate positions $D^{\mathrm{th}}_i$, $D^{\mathrm{st}}_i$ and categories $O_i$ from the $i$-th position head, as illustrated in the Sec.~\ref{sec:position_head}. 
For object centers, we preserve the peak points in $\mathrm{MaxPool}(\mathbf{L}^{\mathrm{th}}_i)$ utilizing a similar method with that in~\cite{zhou2019objects}. 
With the designed kernel fusion and the feature encoder, the prediction $\mathbf{P}$ can be easily produced.
Specifically, we keep the top 100 scoring kernels of objects and all the kernels of stuff after kernel fusion. 
The threshold of 0.4 is utilized to convert predicted soft masks to binary results. 
It should be noted that both the heuristic process or direct $argmax$ could be used to generate non-overlap panoptic results.
The $argmax$ could accelerate the inference but bring a performance drop (1.4\% PQ). 
For fair comparison both from speed and accuracy, the heuristic procedure~\cite{kirillov2019panopticfpn} is adopted in experiments.

\subsection{Point-based Supervision} ~\label{sec:point_supervision}
In the proposed pipeline, the representative kernels are actually located and generated from {\em point-level} object centers and stuff regions in Fig.~\ref{fig:main}, which only requires several points for the kernel generation. 
This motivates us to utilize only a few points as annotation towards weakly-supervised panoptic segmentation that can greatly reduce the human cost for pixel-level annotation. 
To this end, we investigate pointly-supervised manner from three aspects, namely {\em annotation strategy}, {\em target generation}, and {\em data augmentation}.

\subsubsection{Annotation Strategy}
\begin{figure}[!tp]
\centering
\includegraphics[width=0.85\linewidth]{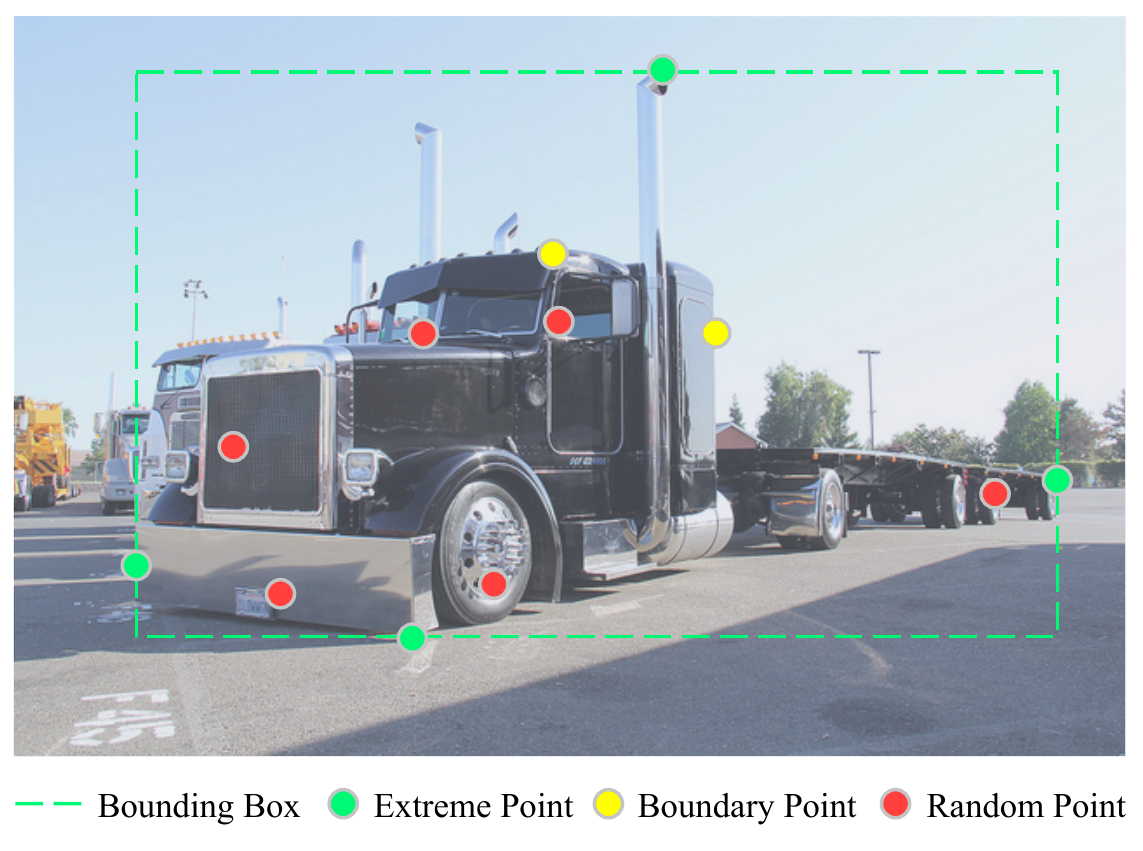} 
\caption{Toy example for different types of point-based annotations. 
We adopt random or boundary points for annotation, regardless of extreme points inside each instance.
}
\label{fig:point_intro}
\end{figure}

For a clear illustration, we first calculate the statistic of {\em boundary} point annotations on COCO {\em training} set in Table~\ref{tab:stat_coco}. 
If all the things and stuff are annotated with boundary points, it respectively requires 25 and 500 points for each object polygon mask and irregular stuff region. 
According to~\cite{lin2014microsoft}, each object instance costs 79s for annotation, and the time consumption for stuff region could reach 45s even with the efficient super-pixel annotation method~\cite{caesar2018coco}.
Because of the limited resource budget in practical scenarios, an efficient strategy to release the potential of annotated points is desired. There are mainly two factors that contribute to the final annotation cost or annotation time, namely point number and point quality. Obviously, more annotated points and better quality enrich the information density but bring much cost. In this way, a simple strategy is randomly choosing $n$ points within each object or stuff region during the annotation period, noted as $\mathcal{P}_n$. 
If 0.9s~\cite{cheng2021pointly} is chosen to measure the annotation cost of each random point, the total cost of $\mathcal{P}_n$ is up to $n\times$0.9s. 
As presented in Table~\ref{tab:stat_coco}, when $\mathcal{P}_{20}$ is adopted for annotation, the point number and time cost are respectively reduced to 10\% and 27\%, while the performance ratio is still maintained as 82\% compared with the fully-annotated version in Table~\ref{tab:abla_point_aug}.
To further improve point quality, a candidate option is choosing points from the boundary of things or stuff, which is investigated in Fig.~\ref{fig:boundary_pq}. 
As depicted in Fig.~\ref{fig:point_intro}, the adopted strategy only requires random sparse points inside each instance for annotation.
Compared with the box-based annotation manner~\cite{cheng2021pointly}, the proposed strategy is unified and can serve both object or stuff region with no need for extreme points. More investigation on annotation strategy is conducted in Sec.~\ref{sec:ablation_point}.

\begin{table}[t!]
\caption{Statistics of {\em boundary} point annotations on COCO {\em training} set. $\mathcal{P}_n$ denotes annotation with $n$ {\em random} points, where the cost of each point is estimated to be 0.9s~\cite{cheng2021pointly}. The cost of super-pixel annotation~\cite{caesar2018coco} is marked as $^\star$.}
\centering
\begin{tabular}{lcccccccccccc}
  \toprule
  Statistics & Point & Instance & Point/Inst & Time/Inst \\
  \midrule
  Thing & 21M  & 850K & 25  & 79s \\
  Stuff & 238M & 475K & 500 & 45s$^\star$ \\
  Total & 259M & 1.3M & 195 & 67s \\
  \midrule
  $\mathcal{P}_{10}$ & 13M & 1.3M & 10 & 9s \\
  $\mathcal{P}_{20}$ & 26M & 1.3M & 20 & 18s \\
  \bottomrule
\end{tabular}
\label{tab:stat_coco}
\end{table}

\begin{figure*}[tp!] 
\centering
\subfigure[Target with convex shape]{
\includegraphics[width=0.3\linewidth]{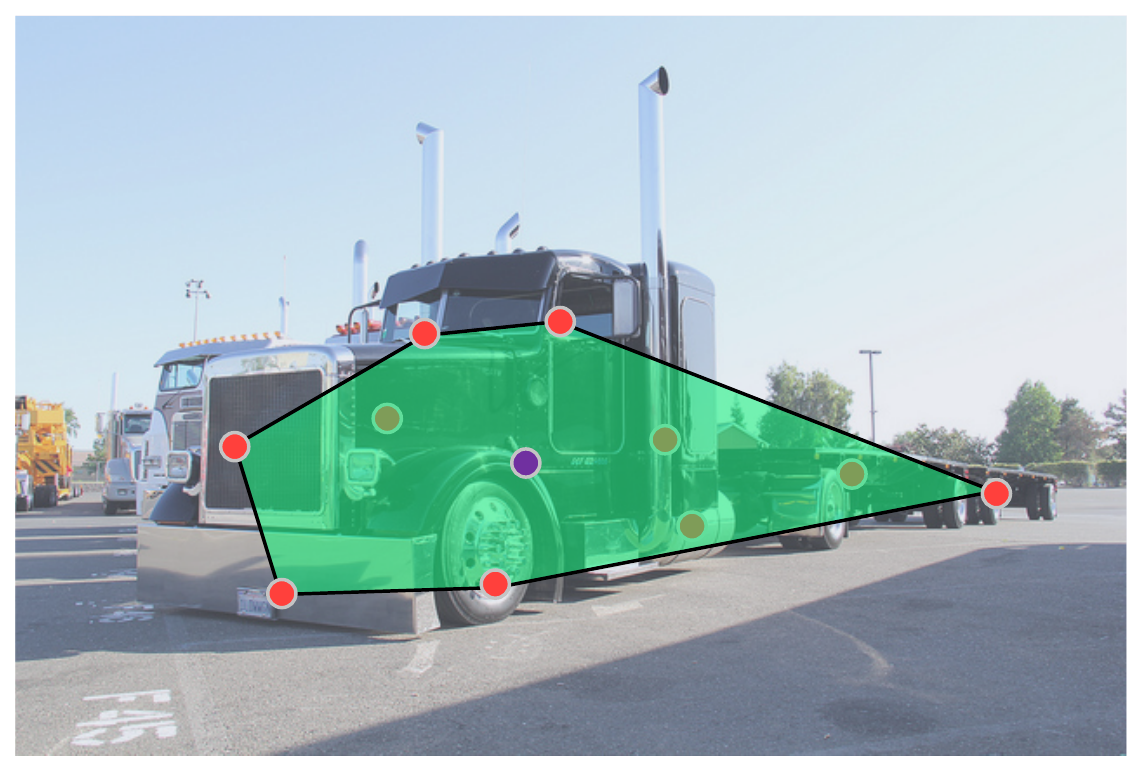}
\label{fig:point_convex}
}
\subfigure[Target with concave shape]{
\includegraphics[width=0.3\linewidth]{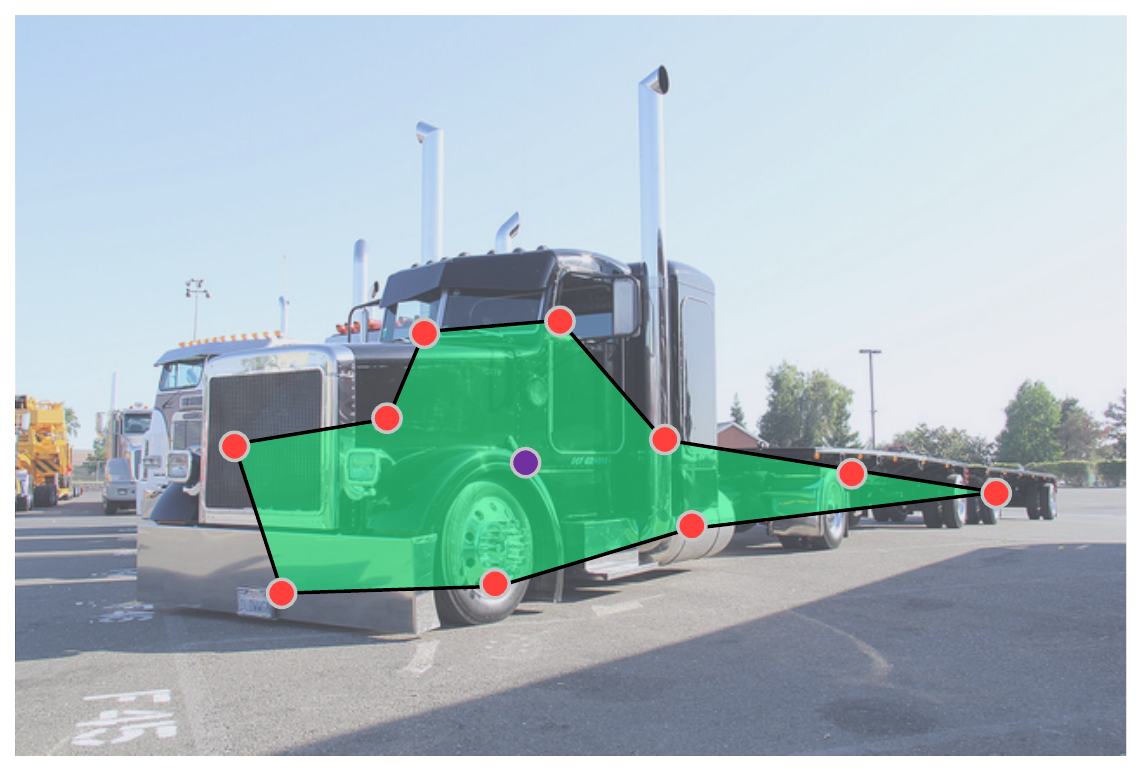}
\label{fig:point_concave}
}
\subfigure[Target for segmentation]{
\includegraphics[width=0.3\linewidth]{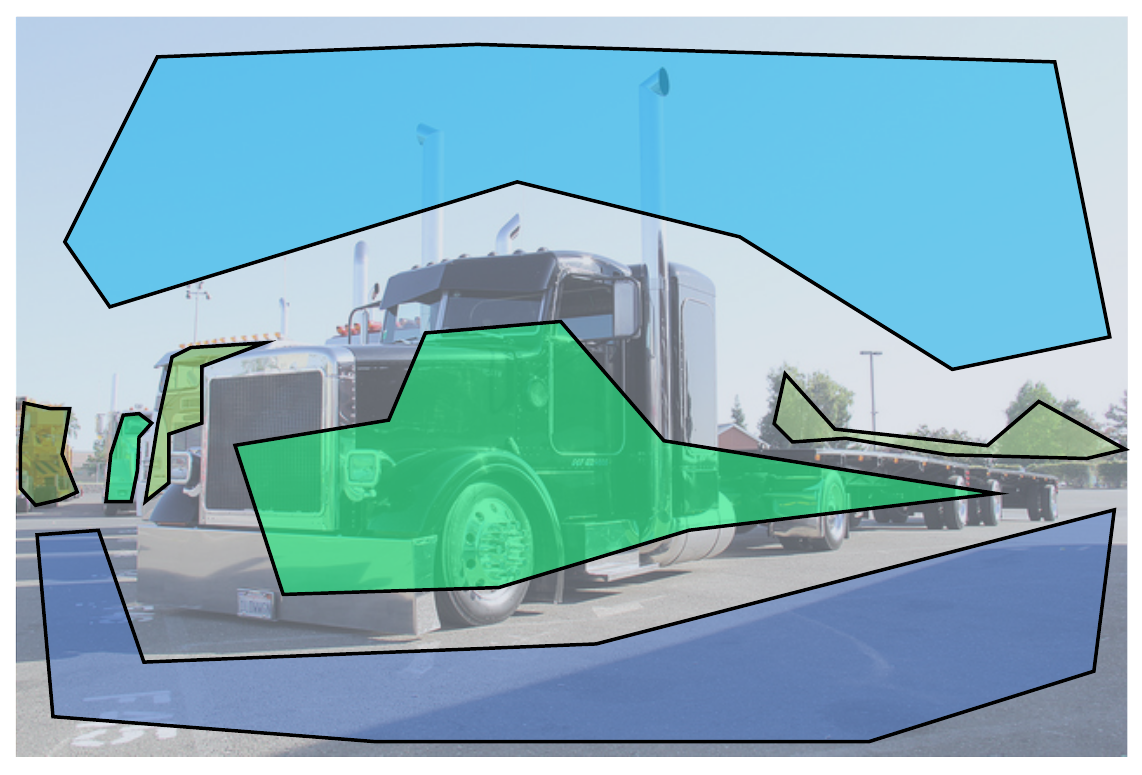}
\label{fig:point_target}
}
\caption{Toy example for $\mathcal{P}_{10}$ target generation. 
The light red points and green region denote the utilized points and the corresponding generated segmentation target, respectively. The dark red points and the purple one represent the unused annotation and the simulated instance center, respectively. Annotated random points in~\ref{fig:point_target} are omitted for concision.}
\vspace{1em}
\label{fig:point_shape}
\end{figure*}

\subsubsection{Target Generation}~\label{sec:target_generation}
Given the annotated points, target is required to optimize ${\mathcal L}_{\mathrm{pos}}$ and ${\mathcal L}_{\mathrm{seg}}$ in the objective function Equ.~\ref{equ:objective_loss}. Recall from Eqs.~\ref{equ:loc} and~\ref{equ:dice_weighted}, supervision for predicted location and segmentation is required to be generated. To this end, we first calculate the mass center from several annotated points as the simulated object center $({\widetilde x}_k, {\widetilde y}_k)$ in Equ.~\ref{equ:thing_gt}. According to the annotation strategy that all the sampled points are located within the instance region, the simulated center is guaranteed to be located around the real object center within that instance. And the experimental results in Sec.~\ref{sec:ablation_point} also prove the effectiveness of this prior. Moreover, to optimize the predicted segmentation $\mathbf{P}$ in Equ.~\ref{equ:dice_weighted}, we generate a binary region for each instance from the given annotated points. As presented in Fig.~\ref{fig:point_shape}, two different strategies are adopted to generate the segmentation target, namely convex shape and concave shape. Compared with convex shape for target generation, the target with concave shape utilizes the given points more adequately and brings finer annotation without introducing negative region, {\em e.g.,} background covered by the convex region in Fig.~\ref{fig:point_convex}. We conduct detailed comparisons between different shapes in Table~\ref{tab:abla_point_num}.

\subsubsection{Data Augmentation}~\label{sec:data_aug}
With the above designed annotation and target generation strategy, the optimization target of the whole image can be easily produced for both things and stuff. To be more specific, we give an example for generated segmentation target with $\mathcal{P}_{10}$ annotation in Fig.~\ref{fig:point_target}. As presented in the diagram, we generate segmentation target with concave shape for each thing or stuff according to the annotated points. The uncovered {\em gray} region is viewed as {\em ignore} during optimization to avoid introducing unreliable target, {\em e.g.,} the background behind the truck. 
For optimization of each specific instance, all the other annotated areas are viewed as negative target to provide pixel-wise supervision. 
Considering the fact that most of the stuff region can not be fully covered with limited random points, we further propose to use shape augmentation to alleviate this side effect. In this paper, we simply adopt the morphological {\em dilation} operator to expand the covered annotated region. Of course, some learnable approaches like generated pseudo-label~\cite{li2018weakly,ahn2019weakly,arun2020weakly} could also be utilized, which will be investigated in the future.
With this method, more background regions can be extended as positive stuff for pixel-level optimization, which is further analyzed in Sec.~\ref{sec:ablation_point}.

\subsubsection{Discussion on Annotation Type}~\label{sec:anno_compare}
The proposed point-based annotation aims to provide supervision for both {\em things} and {\em stuff} in a unified manner.
It differs from previous non-interactive and interactive manners that cannot be directly applied in panoptic segmentation.
In this section, we discuss the connection and difference with related annotation types to make it clear.
We conduct comparisons with related annotation types in Sec.~\ref{sec:ablation_point}.

\vspace{0.5em}
\noindent
\textbf{Non-interactive annotation.}
Non-interactive approaches try to reduce the annotation cost using specially designed types.
Instead of annotating each image pixel-by-pixel, they usually adopt efficient representation to facilitate this process, like point-click~\cite{bearman2016s,cheng2021pointly}, scribble~\cite{lin2016scribblesup}, superpixel~\cite{caesar2018coco}, box~\cite{dai2015boxsup}, and pixel-level block~\cite{lin2019block}.
The proposed point-based annotation is also a type of non-interactive manner and can be integrated with other methods like Block Annotation~\cite{lin2019block}.
But different from previous approaches that are tailored for semantic or instance segmentation, we aim to unify the annotation of things and stuff for panoptic segmentation using several random points.

\vspace{0.5em}
\noindent
\textbf{Interactive annotation.}
Interactive methods are proposed to reduce annotation time with the assistance of pretrained networks.
In this way, users are required to give additional labels for guidance, like drawn box~\cite{rother2004grabcut}, clicked point~\cite{xu2016deep,papadopoulos2017extreme}, and local context~\cite{liew2017regional}.
The interactive manner provides high-quality annotation but needs running models on high-performance machines.
Different from it, the proposed method relies on annotators only without extra cost.
The proposed approach can be further combined with interactive manners to make the annotation more precise.

\vspace{0.5em}
\noindent
\textbf{Point-based annotation.}
Clicking point is a widely-adopted type for efficient annotation.
Generally, point-based method includes several annotation types, {\em e.g.}, extreme points for box~\cite{dai2015boxsup}, sequence of points for scribble~\cite{lin2016scribblesup}, and positive-negative points within box~\cite{cheng2021pointly}.
A most recent approach~\cite{cheng2021pointly} annotates positive and negative points within given boxes.
Such a box-based manner suits object well but can hardly be applied to background stuff for panoptic segmentation.
Therefore, we detach the point-based annotation with predefined boxes and propose to use random points inside each instance for things and stuff simultaneously.

\subsection{Network Details} 
From the aspect of network architecture, ResNet~\cite{he2016deep} with FPN~\cite{lin2017feature} are utilized for backbone instantiation by default. 
We also provide Transformer-based backbones to replace the CNN-based version in experiments. P3 to P7 stages in FPN are used to provide single-stage feature $\mathbf{X}_{i}$ for the kernel generator that is shared across all stages. Meanwhile, P2 to P5 stages are adopted to generate the high-resolution feature $\mathbf{F}^\mathrm{h}$, which is further investigated in Table~\ref{tab:abla_encoder}. 
All convolutions in kernel generator are equipped with $\mathrm{Group Norm}$~\cite{wu2018group} and $\mathrm{ReLU}$ activation. 
Moreover, a naive convolution is adopted at the end of each head in kernel generator for feature projection. 
And a similar strategy with that in \cite{wang2020solov2} is adopted to adjust predicted object scores. 
Our code and models are released to provide more details.

\section{Experiments}~\label{sec:experiments}
In this section, we first introduce experimental settings for Panoptic FCN. Abundant studies are conducted
on the COCO {\em val} set to reveal the effect of each component. 
Finally, comparisons with the previous method are reported.

\subsection{Experimental Setting}

\subsubsection{Datasets} 
We perform extensive experiments on four datasets, namely COCO, VOC 2012, Cityscapes, and Mapillary Vistas.

\vspace{0.5em}
\noindent
\textbf{COCO.} Microsoft COCO~\cite{kirillov2019panoptic,lin2014microsoft,caesar2018coco} is a widely-used benchmark, which contains 80 {\em thing} and 53 {\em stuff} classes. It involves 118K, 5K, and 20K images for training, validation, and testing, respectively. We mainly conduct ablation studies on {\em val} set and report main results on {\em test-dev} set.

\vspace{0.5em}
\noindent
\textbf{VOC 2012.} PASCAL VOC 2012~\cite{everingham2015pascal} contains 20 {\em thing} and one background {\em stuff} category. Following previous work~\cite{li2018weakly,yang2019deeperlab,shen2021toward}, we augment the original {\em training} set with additional annotations from SBD dataset~\cite{BharathICCV2011}, resulting in 10582 images. We report fully- and pointly-supervised results on VOC 2012 {\em validation} set, which contains 1449 images.

\vspace{0.5em}
\noindent
\textbf{Cityscapes.} Cityscapes~\cite{cordts2016cityscapes} consists 5000 street-view {\em fine} annotations with size $1024\times2048$, which are divided into 2975, 500, and 1525 images for training, validation, and testing, respectively. Other {\em coarse} annotations are {\em not} used.

\vspace{0.5em}
\noindent
\textbf{Mapillary Vistas.} Mapillary Vistas~\cite{cordts2016cityscapes} is a large-scale dataset with resolutions from $1024\times768$ to more than $4000\times6000$. 
It includes 37 {\em thing} and 28 {\em stuff} classes with 18K, 2K, and 5K images for training, validation, and testing. 

\subsubsection{Optimization} 
Network optimization is conducted according to different backbones. For CNN-based methods, we use ResNet~\cite{he2016deep} as the backbone and utilize SGD for optimization with learning rate multiplier set to 1. When it comes to Transformer-based approaches, Swin-Transformer~\cite{liu2021Swin} is used as the backbone and AdamW optimizer is adopted with learning rate multiplier set to 0.01. The weight decay and momentum are set to $1e^{-4}$ and 0.9 for both of them. Meanwhile, {\em poly} schedule with power 0.9 is adopted. Experimentally, $\lambda_{\mathrm{pos}}$ is set to a constant 1, and $\lambda_{\mathrm{seg}}$ is respectively set to 3, 3, 4, and 3 for COCO, VOC 2012, Cityscapes, and Mapillary Vistas. 

\subsection{Component-wise Analysis} \label{sec:ablation}
In this section, we give analysis on each component with fully-supervised setting and report results on COCO {\em val} set.

\vspace{0.5em}
\noindent
\textbf{Kernel generator.} Kernel generator plays a vital role in Panoptic FCN and is the key to the {\em instance-awareness} and {\em semantic-consistency}. Here, we compare several settings inside the kernel generator to improve the kernel expressiveness in each stage. As presented in Table~\ref{tab:abla_kernel}, with the number of convolutions in each head increasing, the network performance improves steadily and achieves the peak PQ with 3 stacked $\mathrm{Conv3\times3}$ whose channel number is 256. Additional operations in the head brings no more gain. Similar with~\cite{zhou2019objects}, deformable convolutions~\cite{zhu2019deformable} are adopted in position head to extend the receptive field, which brings further improvement, especially in stuff regions (1.4\% PQ). 

\vspace{0.5em}
\noindent
\textbf{Position embedding.} Due to the {\em instance-aware} property of objects, position embedding is introduced to provide essential cues. In Table~\ref{tab:abla_position}, we compare among several positional settings by attaching relative coordinates~\cite{liu2018intriguing} to different heads. An interesting finding is that the improvement is minor (up to 0.3\% PQ) if coordinates are attached to the kernel head or feature encoder only, but it boosts to 1.4\% PQ when given the positional cues to both heads. It can be attributed to the constructed correspondence in the position between kernel weights and the encoded feature.

\vspace{0.5em}
\noindent
\textbf{Kernel fusion.} Kernel fusion is a core operation in the proposed method, which guarantees the required properties for things and stuff, as elaborated in Sec.~\ref{sec:kernel_fusion}. We investigate the fusion type {\em class-aware} and similarity thresholds {\em thres} in Table~\ref{tab:abla_similar}. As shown in the table, the network attains the best performance 41.3\% PQ with {\em thres} 0.90. When we decrease the {\em thres}, more kernels with lower similarity are merged to predict the identical instance, which brings inferior performance. And the {\em class-agnostic} manner could dismiss some similar instances with different categories, which yields little drop in PQ result. 

\begin{table}[t!]
\caption{Comparison with different settings of kernel generator on the COCO {\em val} set. {\em deform} and {\em conv num} denote deformable convolutions and number of convolutions.}
\centering
\begin{tabular}{ccccccc}
  \toprule
  {\em {deform}} & {\em{conv num}} & PQ & PQ$^\mathrm{th}$ & PQ$^\mathrm{st}$ \\ 
  \midrule
  \xmark & 1 & 38.4 & 43.4 & 31.0 \\ 
  \xmark & 2 & 38.9 & 44.1 & 31.1 \\ 
  \xmark & 3 & 39.2 & 44.7 & 31.0 \\ 
  \xmark & 4 & 39.2 & 44.9 & 30.8 \\ 
  \midrule
  \cmark & 3 & {\bf 39.9} & {\bf 45.0} & {\bf 32.4} \\
  \bottomrule
\end{tabular}
\label{tab:abla_kernel}
\end{table}

\begin{table}[t!]
\caption{Comparison with different positional settings on the COCO {\em val} set. {\em coord}$_\mathrm{w}$ and {\em coord}$_\mathrm{f}$ denote combining coordinates for the kernel head, and feature encoder.}
\centering
\begin{tabular}{ccccccc}
  \toprule
  {\em coord}$_\mathrm{w}$ & {\em coord}$_\mathrm{f}$ & PQ & PQ$^\mathrm{th}$ & PQ$^\mathrm{st}$ \\ 
  \midrule
  \xmark & \xmark & 39.9 & 45.0 & 32.4 \\ 
  \midrule
  \cmark & \xmark & 39.9 & 45.0 & 32.2 \\ 
  \xmark & \cmark & 40.2 & 45.3 & 32.5 \\ 
  \cmark & \cmark & {\bf 41.3} & {\bf 46.9} & {\bf 32.9} \\ 
  \bottomrule
\end{tabular}
 \label{tab:abla_position}
\end{table}

\begin{table}[t!]
 \caption{Comparison with different similarity thresholds of kernel fusion on the COCO {\em val} set. {\em{class-aware}} denotes merging kernel weights with the same predicted class. {\em thres} indicates the cosine similarity threshold for kernel fusion.}
 \centering
\begin{tabular}{ccccccc}
  \toprule
   {\em{class-aware}} & {\em{thres}} & PQ & PQ$^\mathrm{th}$ & PQ$^\mathrm{st}$ \\ 
  \midrule
  \cmark & 0.80 & 39.7 & 44.3 & 32.9 \\ 
  \cmark & 0.85 & 40.8 & 46.1 & 32.9 \\ 
  \cmark & 0.90 & {\bf 41.3} & 46.9 & {\bf 32.9} \\ 
  \cmark & 0.95 & 41.3 & {\bf 47.0} & 32.9 \\ 
  \cmark & 1.00 & 38.7 & 42.6 & 32.9 \\ 
  \midrule
  \xmark & 0.90 & 41.2 & 46.7 & 32.9 \\ 
  \bottomrule
\end{tabular}
 \label{tab:abla_similar}
\end{table}

\begin{table}[t!]
 \caption{Comparison with different methods of removing repetitive predictions. {\em kernel-fusion} and {\em nms} indicates the proposed kernel-level fusion method and Matrix NMS~\cite{wang2020solov2}.}
 \centering
\begin{tabular}{ccccccc}
\toprule
   {\em{kernel-fusion}} & {\em{nms}} & PQ & PQ$^\mathrm{th}$ & PQ$^\mathrm{st}$ \\ 
   \midrule
   \xmark & \xmark & 38.7 & 42.6 & 32.9 \\ 
   \xmark & \cmark & 38.7 & 42.6 & 32.9 \\ 
   \cmark & \xmark & {\bf 41.3} & {\bf 46.9} & {\bf 32.9} \\ 
   \cmark & \cmark & 41.3 & 46.9 & 32.8 \\
  \bottomrule
\end{tabular}
 \label{tab:abla_nms}
\end{table}

\begin{table}[t!]
 \caption{Comparison with different channel numbers of the feature encoder on the COCO {\em val} set. {\em channel num} represents the channel number $C_\mathrm{e}$ of the feature encoder.}
 \centering
\begin{tabular}{cccccc}
  \toprule
  {\small \em{channel num}} & PQ & PQ$^\mathrm{th}$ & PQ$^\mathrm{st}$ \\
  \midrule
  16 & 39.9 & 45.0 & 32.1 \\
  32 & 40.8 & 46.3 & 32.5 \\
  64 & {\bf 41.3} & 46.9 & {\bf 32.9} \\
  128 & 41.3 & {\bf 47.0} & 32.6 \\
  \bottomrule
\end{tabular}
 \label{tab:abla_channel}
\end{table}

\begin{table}[t!]
 \caption{Comparison with different feature types for the feature encoder on the COCO {\em val} set. {\em feature type} denotes the method to generate high-resolution feature $\mathbf{F}^{\mathrm{h}}$ in Sec.~\ref{sec:feature_encoder}.}
 \centering
\begin{tabular}{lccccc}
  \toprule
  {\em{feature type}} & PQ & PQ$^\mathrm{th}$ & PQ$^\mathrm{st}$ \\ 
  \midrule
  FPN-P2 & 40.6 & 46.0 & 32.4 \\ 
  FPN-Summed & 40.5 & 46.0 & 32.1 \\
  Semantic FPN~\cite{kirillov2019panopticfpn} & {\bf 41.3} & {\bf 46.9} & {\bf 32.9} \\ 
  \bottomrule
\end{tabular}
\label{tab:abla_encoder}
\end{table}

\begin{table}[t!]
 \caption{Comparison with different settings of weighted dice loss on the COCO {\em val} set. {\em weighted} and $k$ denote weighted dice loss and the sampled number in Sec.~\ref{sec:train_infer}.}
 \centering
\begin{tabular}{ccccccc}
  \toprule
  {\small \em weighted } & $k$ & PQ & PQ$^\mathrm{th}$ & PQ$^\mathrm{st}$ \\ 
  \midrule
  \xmark & - & 40.2 & 45.5 & 32.4 \\ 
  \midrule
  \cmark & 1 & 40.0 & 45.1 & 32.4 \\ 
  \cmark & 3 & 41.0 & 46.4 & 32.7 \\ 
  \cmark & 5 & 41.0 & 46.5 & 32.9 \\ 
  \cmark & 7 & {\bf 41.3} & {\bf 46.9} & {\bf 32.9} \\ 
  \cmark & 9 & 41.3 & 46.8 & 32.9 \\ 
  \bottomrule
\end{tabular}
 \label{tab:abla_dice}
\end{table}

\vspace{0.5em}
\noindent
\textbf{Comparison with NMS.} Furthermore, we compare kernel fusion with Matrix NMS~\cite{wang2020solov2} that is utilized for pixel-level instance removal in post-processing. As presented in Table~\ref{tab:abla_nms}, the proposed method works well and the performance saturates with the simple kernel-level fusion, which improves 2.6\% PQ over the baseline. And the extra NMS brings no more gain to the performance.


\vspace{0.5em}
\noindent
\textbf{Feature encoder.} To enhance expressiveness of the encoded feature $\mathbf{F}^\mathrm{e}$, we further explore the {\em channel number} and {\em feature type} used in feature encoder. As illustrated in Table~\ref{tab:abla_channel}, the network achieves 41.3\% PQ with 64 channels, and extra channels contribute little improvement. For efficiency, we set the channel number of feature encoder to 64 by default. As for high-resolution feature generation, three types of methods are further discussed in Table~\ref{tab:abla_encoder}. Here, FPN-P2 denotes using feature from P2 stage of FPN directly, and FPN-Summed represents using the summation of rescaled feature from all the FPN stages. It is clear that Semantic FPN~\cite{kirillov2019panopticfpn}, which combines features from four stages in FPN step-by-step, achieves the top performance 41.3\% PQ.

\vspace{0.5em}
\noindent
\textbf{Weighted dice loss.} The designed weighted dice loss aims to release the potential of kernel generator by sampling $k$ positive kernels inside each object. Compared with the original dice loss, which selects a single central point in each object, improvement brought by the weighted dice loss reaches 1.1\% PQ, as presented in Table~\ref{tab:abla_dice}. This is achieved by sampling 7 top-scoring kernels to generate results of each instance, which are optimized together in each step. And the performance saturates with 7 samples.

\vspace{0.5em}
\noindent
\textbf{Center type.} To better optimize the position head, we have explored different types of center assigning strategies. For object center generation, {\em mass center} is utilized to provide the $k$-th ground-truth coordinate $\widetilde x_k$ and $\widetilde y_k$. As presented in Table~\ref{tab:abla_centertype}, compared with the {\em box center} for ground-truth, we find that {\em mass center} brings superior performance and higher robustness. It can be attributed to that most of the mass centers are located within the object area, while it is not the case for box centers.

\vspace{0.5em}
\noindent
\textbf{Enhanced version.} 
We explore model capacity by combining existing simple enhancements in feature encoder, {\em e.g.,} deformable convolutions and extra channels. As illustrated in Table~\ref{tab:abla_enhance}, the reinforcement contributes 0.7\% improvement over the default setting, marked as {\bf Panoptic FCN*}. 

\vspace{0.5em}
\noindent
\textbf{Upper-bound analysis.} 
In Table~\ref{tab:abla_upperbound}, we give analysis to the upper bound of the designed fashion with Res50-FPN backbone. 
As shown in the table, given ground-truth positions of $\mathbf{L}_i^{\mathrm{th}}$ and $\mathbf{L}_i^{\mathrm{st}}$, the network yields 6.2\% PQ. And it brings extra boost ({\bf 16.1}\% PQ) to the network if we assign ground-truth categories to the position head. Compared with the baseline, there still remains huge potential to be explored, especially for stuff which has even up to {\bf 33.7}\% PQ.

\begin{figure}[t!]
\centering
\includegraphics[width=\linewidth]{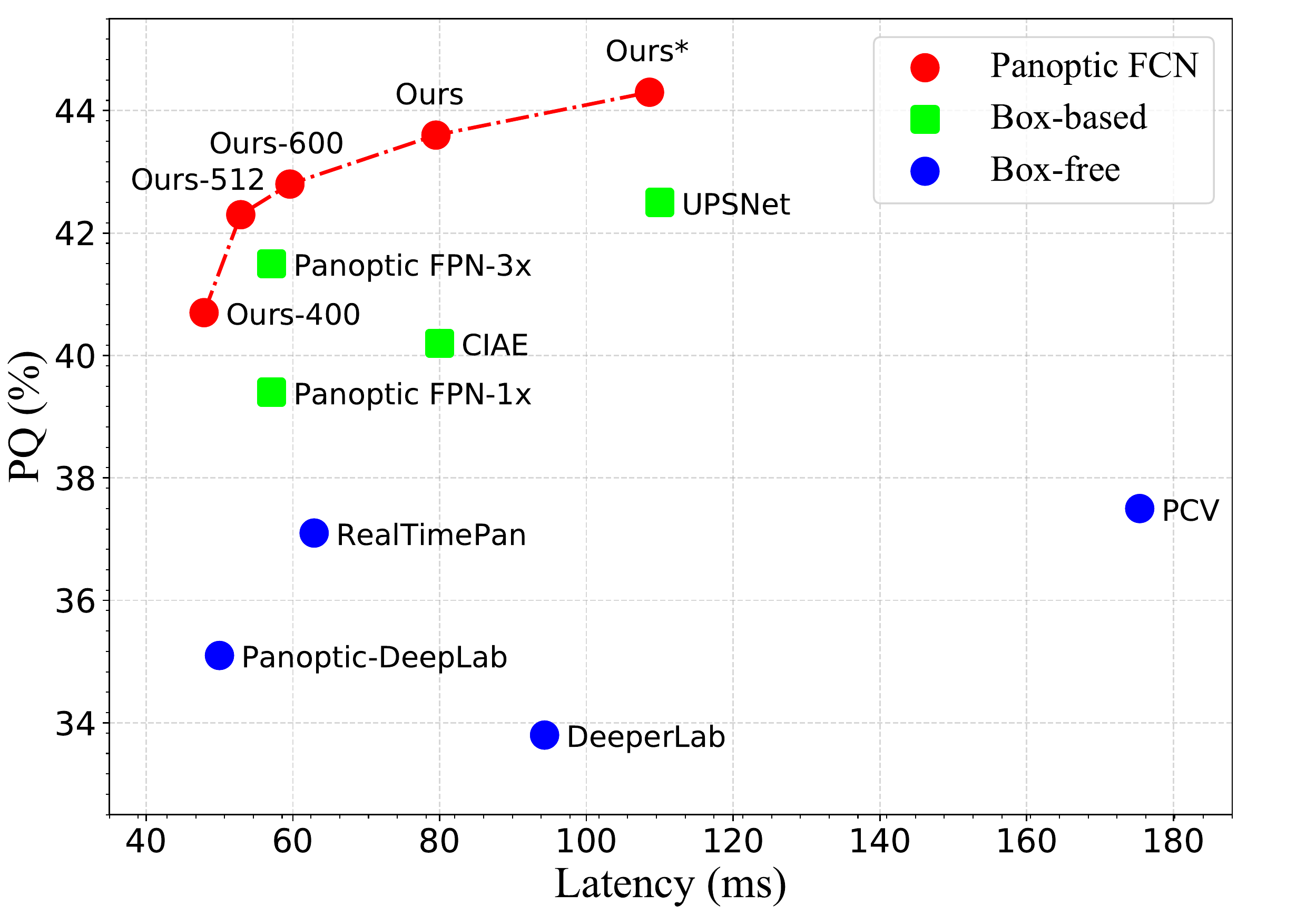} 
\caption{Speed-Accuracy trade-off curve on the COCO {\em val} set. 
Results are compared with Res50 except DeeperLab~\cite{yang2019deeperlab} with Xception-71~\cite{chollet2017xception}. The latency is measured {\em end-to-end} from single input to panoptic result. Details are given in Table~\ref{tab:coco_val}.
}

\label{fig:speed_acc}
\end{figure}

\vspace{0.5em}
\noindent
\textbf{Speed-accuracy.} 
To verify the network efficiency, we plot the speed-accuracy trade-off curve on the COCO {\em val} set. 
As presented in Fig.~\ref{fig:speed_acc}, Panoptic FCN surpasses all previous {\em box-free} models by large margins on both performance and efficiency.
Even compared with the well-optimized Panoptic FPN~\cite{kirillov2019panopticfpn} from $\mathrm{Detectron2}$~\cite{wu2019detectron2}, our approach still attains a better speed-accuracy balance with different image scales. 

\begin{table}[t!]
\centering
 \caption{Comparisons with different settings of center type on the COCO {\em val} set. 
 {\em weighted} and {\em center type} denote weighted dice loss and center type for ground-truth.}
\begin{tabular}{ccccccc}
  \toprule
  {\small \em weighted }& {\small \em center type} & PQ & PQ$^\mathrm{th}$ & PQ$^\mathrm{st}$ \\
  \midrule
  \xmark & box & 39.7 & 44.7 & 32.3 \\
  \xmark & mass & 40.2 & 45.5 & 32.4 \\
  \midrule
  \cmark & box & 40.6 & 45.7 & 32.8 \\ 
  \cmark & mass & {\bf 41.3} & {\bf 46.9} & {\bf 32.9} \\
  \bottomrule
\end{tabular}
 \label{tab:abla_centertype}
\end{table}

\begin{table}[t!]
 \caption{Comparison with different settings of the feature encoder on the COCO {\em val} set. 
 {\em deform} and {\em channel num} denote deformable convolutions and the channel number $C_\mathrm{e}$.}
 \centering
\begin{tabular}{ccccccc}
  \toprule
  {\em{deform}} & {\em{channel num}} & PQ & PQ$^\mathrm{th}$ & PQ$^\mathrm{st}$ \\
  \midrule
  \xmark & 64  & 43.6 & 49.3 & 35.0 \\ 
  \cmark & 256 & {\bf 44.3} & {\bf 50.0} & {\bf 35.6} \\ 
  \bottomrule
\end{tabular}
 \label{tab:abla_enhance}
\end{table}

\begin{table}[t!]
 \caption{Upper-bound analysis on the COCO {\em val} set. 
 {\em{gt position}} and {\em{gt class}} denote utilizing the ground-truth position $G_i$ and class $O_i$ in each position head.}
 \centering
\begin{tabular}{ccccccc}
  \toprule
  {\em{gt position}} & {\em{gt class}} & PQ & PQ$^\mathrm{th}$ & PQ$^\mathrm{st}$ \\ 
  \midrule
  \xmark & \xmark & 43.6 & 49.3 & 35.0 \\ 
  \cmark & \xmark & 49.8 & 52.2 & 46.1 \\
  \cmark & \cmark & {\bf 65.9} & {\bf 64.1} & {\bf 68.7} \\ 
  \midrule
    &  & {\em +22.3} & {\em +14.8} & {\em +33.7} \\
  \bottomrule
\end{tabular}
 \label{tab:abla_upperbound}
\end{table}

\begin{table}[t!]
\centering
 \caption{Stability analysis of point-based annotation on the COCO {\em val} set. We calculate point-based annotations for 5 times with different random seeds in {\em test} experiments.}
\begin{tabular}{lccccccc}
  \toprule
  Statistics & {\em test-1} & {\em test-2} & {\em test-3} & {\em test-4} & {\em test-5} & {\em mean} & {\em std}\\
  \midrule
  PQ & 34.0 & 34.0 & 34.0 & 34.0 & 33.8 & 33.96 & 0.08 \\
  PQ$^\mathrm{th}$ & 39.0 & 39.1 & 39.1 & 39.1 & 38.8 & 39.01 & 0.12 \\
  PQ$^\mathrm{st}$ & 26.5 & 26.4 & 26.3 & 26.4 & 26.3 & 26.38 & 0.07 \\
  \bottomrule
\end{tabular}
 \label{tab:abla_stability}
\end{table}

\subsection{Pointly-supervised Analysis} \label{sec:ablation_point}
We give analysis to the point-based annotation from annotation stability, point number, target shape, boundary point ratio, shape augmentation, and annotation type.

\vspace{0.5em}
\noindent
\textbf{Annotation stability.} 
To ensure stability of the point-based annotation, we validate it by sampling random points from COCO {\em training} set for 5 times and generate point-based targets. 
In particular, we adopt random seeds and write the sampled points to a fix file each time. 
Then we generate simulated centers and concave segmentation targets following the designed strategy in Sec.~\ref{sec:point_supervision}. 
With the generated target, we optimize the network with $\mathrm{1\times}$ schedule and report the performance in Table~\ref{tab:abla_stability}. 
The performance with different sampling points proves the stability of pointly-supervised network, where the standard deviation of PQ is only 0.08.

\begin{figure}[t!]
\centering
\includegraphics[width=\linewidth]{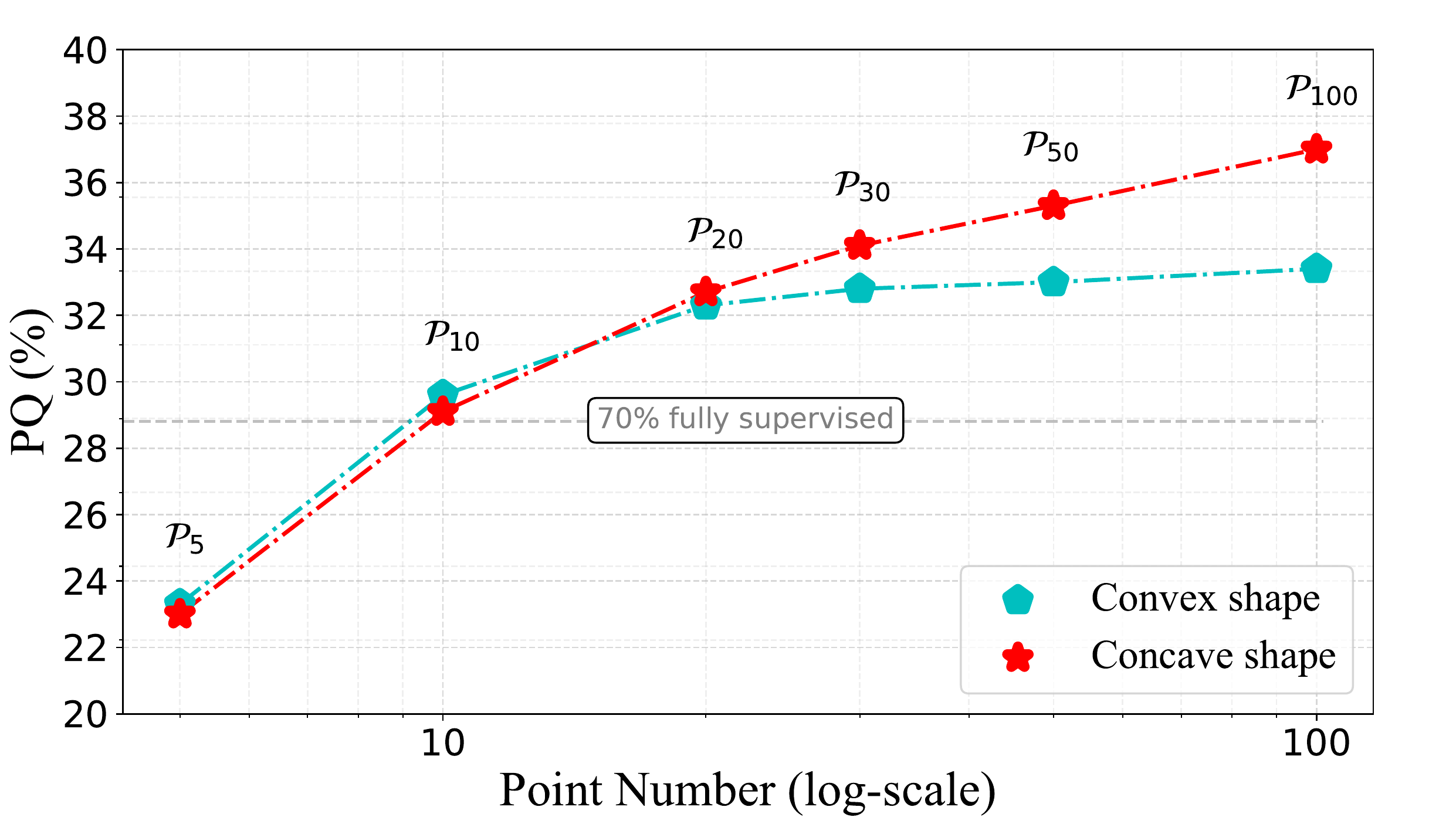} 
\caption{Curve of point number on the COCO {\em val} set with $\mathrm{1\times}$ training schedule. The network outperforms 70\% of the fully-supervised version with point-based annotation $\mathcal{P}_{10}$.}
\label{fig:point_pq}
\end{figure}

\vspace{0.5em}
\noindent
\textbf{Point number.}
Traditional fully annotated masks $\mathcal{M}$ usually require a huge amount of points for arbitrary shape, as illustrated in Table~\ref{tab:stat_coco}. To achieve cost-accuracy trade-off, we investigate the relationship between annotated point number and performance in Table~\ref{tab:abla_point_num} and Fig.~\ref{fig:point_pq}. As presented in the diagram, more points bring better performance from $\mathcal{P}_{5}$ to $\mathcal{P}_{100}$, which matches our prior knowledge well that finer annotation contributes higher result. 
And the pointly-supervised network with $\mathcal{P}_{10}$ exceeds 70\% performance of the fully-supervised version, which only accounts for 13\% annotation cost of fully-annotated mask $\mathcal{M}$.

\vspace{0.5em}
\noindent
\textbf{Target shape.}
Shape of the generated target plays a vital role in high-quality segmentation. 
Different target shapes proposed in Sec.~\ref{sec:target_generation} are compared in Table~\ref{tab:abla_point_num} and Fig.~\ref{fig:point_pq}. 
It is apparent that the performance of convex shape surpasses that of concave shape with a handful of annotated points but is inferior to the concave one if given sufficient points. 
This can be attributed to that the convex target usually covers more area than concave shape, but it introduces negative region with more annotated points, as shown in Fig.~\ref{fig:point_shape}. 
Consequently, the performance with convex target saturates with annotated points increase to $\mathcal{P}_{20}$, while it is not the case for concave shape.  And the point-based annotation is set to $\mathcal{P}_{20}$ by default for subsequent experiments.

\begin{table}[t!]
 \caption{Comparisons with different supervisions on the COCO {\em val} set. 
 {\em shape} and {\em{supervision}} denote using different shape and supervised strategy. 
 Percentage indicates PQ ratio of pointly-supervised $\mathcal{P}$ in mask-supervised $\mathcal{M}$.}
 \centering
\begin{tabular}{cccccc}
  \toprule
  {\em{shape}} & {\em{supervision}} & PQ & PQ$^\mathrm{th}$ & PQ$^\mathrm{st}$ & Percentage\\
  \midrule
  \multirow{6}{*}{convex} & $\mathcal{P}_{5}$ & 23.3 & 29.3 & 14.3 & 56.4 \\
    & $\mathcal{P}_{10}$ & 29.6 & 35.3 & 21.0 & 71.6 \\
    & $\mathcal{P}_{20}$ & 32.3 & 37.5 & 24.4 & 78.2 \\
    & $\mathcal{P}_{30}$ & 32.8 & 37.7 & 25.2 & 79.4 \\
    & $\mathcal{P}_{50}$ & 33.0 & 37.8 & 25.7 & 79.9 \\
    & $\mathcal{P}_{100}$ & 33.4 & 38.3 & 26.0 & 80.8 \\
  \midrule
  \multirow{6}{*}{concave} & $\mathcal{P}_{5}$ & 23.0 & 28.7 & 14.3 & 55.6 \\
    & $\mathcal{P}_{10}$ & 29.1 & 35.0 & 20.2 & 70.4 \\
    & $\mathcal{P}_{20}$ & 32.7 & 38.1 & 24.5 & 79.1 \\
    & $\mathcal{P}_{30}$ & 34.1 & 39.4 & 26.2 & 82.5 \\
    & $\mathcal{P}_{50}$ & 35.3 & 40.2 & 27.9 & 85.4 \\
    & $\mathcal{P}_{100}$ & 37.0 & 41.8 & 29.6 & 89.5 \\
  \midrule
  concave & $\mathcal{M}$ & {\bf 41.3} & {\bf 46.9} & {\bf 32.9} & {\bf 100.0} \\
  \bottomrule
\end{tabular}
 \label{tab:abla_point_num}
\end{table}

\begin{figure}[t!]
\centering
\includegraphics[width=\linewidth]{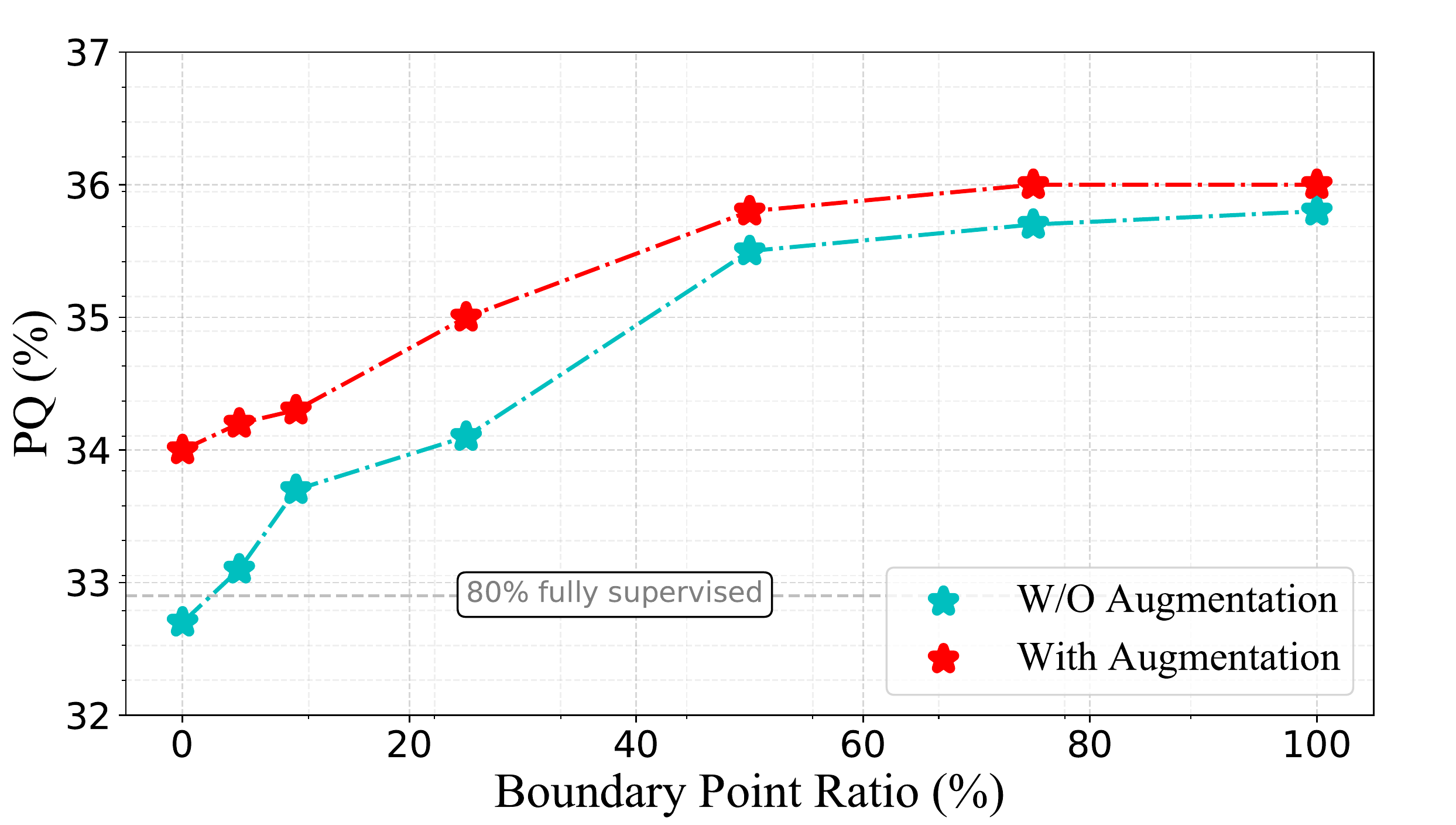} 
\caption{Curve of boundary point ratio on the COCO {\em val} set with $\mathrm{1\times}$ schedule and point-based annotation $\mathcal{P}_{20}$. 
The network outperforms 80\% of the fully supervised version.}
\label{fig:boundary_pq}
\end{figure}

\vspace{0.5em}
\noindent
\textbf{Boundary point ratio.}
As declared in Fig.~\ref{fig:point_intro}, boundary points describe the maximum positive region of things or stuff, which is another crucial factor for target quality. And the point-based annotation $\mathcal{P}$ actually turns to fully-annotated mask $\mathcal{M}$ if all the boundary points are given. We compare with different ratios of boundary points based on $\mathcal{P}_{20}$ annotation, as presented in Table~\ref{tab:abla_point_aug} and Fig.~\ref{fig:boundary_pq}. It is obvious that the performance increases with more boundary points and saturates when the ratio reaches 50\%.

\vspace{0.5em}
\noindent
\textbf{Shape augmentation.}
We conduct ablation studies on the dilated shape augmentation in Sec.~\ref{sec:data_aug}. 
As shown in Table~\ref{tab:abla_point_aug} and Fig.~\ref{fig:boundary_pq}, the augmentation strategy improves steadily with more boundary point ratios. 
In particular, it contributes 1.3\% PQ gain over the baseline with no boundary point, and achieves 82.3\% of the fully-supervised version. 
The effect of shape augmentation decreases with more boundary points that bring sufficient positive regions.

\begin{table}[t!]
 \caption{Comparisons with different strategies of $\mathcal{P}_{20}$ on the COCO {\em val} set.
 {\em shape aug} and {\em point ratio} denote shape augmentation and boundary-point ratio. 
 Percentage indicates PQ ratio of pointly-supervised $\mathcal{P}$ in mask-supervised $\mathcal{M}$.}
 \centering
\begin{tabular}{cccccc}
  \toprule
  {\em{shape aug}} & {\em{point ratio}} & PQ & PQ$^\mathrm{th}$ & PQ$^\mathrm{st}$ & Percentage\\
  \midrule
  \multirow{7}{*}{\xmark} & 0\% & 32.7 & 38.1 & 24.5 & 79.1 \\
    & 5\% & 33.1 & 38.6 & 24.8 & 80.1 \\
    & 10\% & 33.7 & 39.1 & 25.5 & 81.5 \\
    & 25\% & 34.1 & 39.3 & 26.4 & 82.5 \\
    & 50\% & 35.5 & 41.0 & 27.2 & 85.9 \\
    & 75\% & 35.7 & 41.2 & 27.5 & 86.4 \\
    & 100\% & 35.8 & 41.2 & 27.6 & 86.6 \\
  \midrule
  \multirow{7}{*}{\cmark} & 0\% & 34.0 & 39.0 & 26.4 & 82.3 \\
    & 5\% & 34.2 & 39.3 & 26.6 & 82.8 \\
    & 10\% & 34.3 & 39.3 & 26.8 & 83.0 \\
    & 25\% & 35.0 & 40.2 & 27.1 & 84.7 \\
    & 50\% & 35.8 & 41.2 & 27.6 & 86.6 \\
    & 75\% & 36.0 & 41.4 & 27.8 & 87.1 \\
    & 100\% & 36.0 & 41.4 & 27.8 & 87.1 \\
  \midrule
  \xmark & 100\% & {\bf 41.3} & {\bf 46.9} & {\bf 32.9} & {\bf 100.0} \\
  \bottomrule
\end{tabular}
 \label{tab:abla_point_aug}
\end{table}

\begin{table}[t!]
 \caption{Comparison with different training schedules on the COCO {\em val} set. PQ$_{\mathcal{P}_{n}}$ indicates the performance with $\mathcal{P}_{n}$ supervision. $\mathrm{1\times}$, $\mathrm{2\times}$, and $\mathrm{3\times}$ {\em schedule} denote the $\mathrm{90K}$, $\mathrm{180K}$, and  $\mathrm{270K}$ training iterations in $\mathrm{Detectron2}$~\cite{wu2019detectron2}.}
 \centering
\begin{tabular}{cccccccc}
  \toprule
  {\small \em{schedule}} & PQ$_{\mathcal{P}_{20}}$ & PQ$_{\mathcal{P}_{30}}$ & PQ & PQ$^\mathrm{th}$ & PQ$^\mathrm{st}$  \\
  \midrule
  $\mathrm{1\times}$ & 34.0 & 35.0 & 41.3 & 46.9 & 32.9 \\ 
  $\mathrm{2\times}$ & 35.0 & 36.6 & 43.2 & 48.8 & 34.7 \\
  $\mathrm{3\times}$ & {\bf 35.1} & {\bf 36.6} & {\bf 43.6} & {\bf 49.3} & {\bf 35.0} \\
  \bottomrule
\end{tabular}
 \label{tab:abla_training}
\end{table}

\vspace{0.5em}
\noindent
\textbf{Training schedule.} 
To fully optimize the network, we prolong the training iteration to the $3\times$ training schedule, which is widely adopted in recent one-stage instance-level approaches~\cite{chen2019tensormask,wang2019solo,wang2020solov2}. As shown in Table~\ref{tab:abla_training}, for pointly-supervised network, $2\times$ training schedule respectively brings 1.0\% and 1.6\% PQ improvements for $\mathcal{P}_{20}$ and $\mathcal{P}_{30}$. And increasing iterations to $3\times$ schedule contributes very little. 
As for fully-supervised version, $2\times$ training schedule contributes 1.9\% PQ gain and more iterations improve extra 0.4\% PQ. The relative improvement gap between different supervisions reflects that optimization improvement is in proportion to the annotation quality, {\em e.g.,} the annotated point number.

\vspace{0.5em}
\noindent
\textbf{Annotation type.}
In Table~\ref{tab:abla_anno}, we compare with related annotation types for semantic segmentation on the VOC 2012 dataset.
In particular, we keep the semantic meaning of the annotated instances and drop the head for object center prediction in Fig.~\ref{fig:main}.
Panoptic FCN with Res50 achieves 62.2\% mIoU based on $\mathcal{P}_{10}$ annotation.
The performance steadily improves to 64.6\% and 66.4\% mIoU with 20 and 30 random points for each instance, which surpasses other point-based and box-based methods with a large margin.
We further conduct detailed comparisons with non-interactive~\cite{lin2019block} and interactive~\cite{maninis2018deep} annotations in Table~\ref{tab:abla_anno_detail}.
Compared with BlackAnno~\cite{lin2019block} that annotates pixel-level blocks, the point-based annotation with $\mathcal{P}_{10}$ attains similar performance using less time and a much reduced annotated pixel ratio.
As for DEXTR~\cite{maninis2018deep} that utilizes an interactive manner to generate annotation, the proposed approach achieves better results and less time cost based on $\mathcal{P}_{30}$ annotation.
Overall, the proposed manner achieves 85\% of the fully-supervised performance with 18\% time cost based on $\mathcal{P}_{20}$ annotation.
The performance ratio with $\mathcal{P}_{30}$ is further improved to 88\% if time is released to 26\% of the fully-supervised version.

\begin{table}[t!]
 \caption{
 Comparison with related annotation types on the VOC 2012 {\em val} set.
 $\mathcal{S}$, $\mathcal{B}$, $\mathcal{I}$, and $\mathcal{P}$ denote scribble, box-, image-, and point-level annotation, respectively.
}
 \centering
\begin{tabular}{lccccccc}
  \toprule
  Method & Supervision & mIoU  \\ 
  \midrule
  MIL-FCN~\cite{pathak2014fully} & $\mathcal{I}$ & 25.1 \\ 
  WSSL~\cite{papandreou2015weakly} & $\mathcal{I}$ & 38.2 \\
  point sup.~\cite{bearman2016s} & $\mathcal{P}$ & 46.1 \\ 
  ScribbleSup~\cite{lin2016scribblesup} & $\mathcal{P}$ & 51.6 \\
  WSSL~\cite{papandreou2015weakly} & $\mathcal{B}$ & 60.6 \\
  BoxSup~\cite{dai2015boxsup} & $\mathcal{B}$ & 62.0 \\
  ScribbleSup~\cite{lin2016scribblesup} & $\mathcal{S}$ & 63.1 \\
  \midrule
  PanopticFCN & $\mathcal{P}_{10}$ & 62.2 \\
  PanopticFCN & $\mathcal{P}_{20}$ & 64.6 \\
  PanopticFCN & $\mathcal{P}_{30}$ & {\bf 66.4} \\
  \bottomrule
\end{tabular}
 \label{tab:abla_anno}
\end{table}

\begin{table}[t!]
 \caption{
 Detailed comparisons with related types on the VOC 2012 {\em val} set.
 $\mathcal{A}$, $\mathcal{M}$, and $\mathcal{P}$ denote automatic, mask-, and point-level annotation.
 Models are pretrained on COCO dataset.
 Only DEXTR~\cite{maninis2018deep} adopts multi-scale testing.
 Ratio and Time indicate the annotated pixel ratio and cost time.
}
 \centering
\begin{tabular}{lccccccc}
  \toprule
  Method & Supervision & Ratio & Time & mIoU  \\ 
  \midrule
  BlockAnno~\cite{lin2019block} & $\mathcal{M}$ & 10\% & 25s & 67.2 \\
  DEXTR~\cite{maninis2018deep} & $\mathcal{P}$+$\mathcal{A}$ & - & 420s & 70.0 \\
  \midrule
  PanopticFCN & $\mathcal{P}_{10}$ & $<$1\% & 21s & 66.3 \\
  PanopticFCN & $\mathcal{P}_{20}$ & $<$1\% & 42s & 68.4 \\
  PanopticFCN & $\mathcal{P}_{30}$ & $<$1\% & 63s & 70.9 \\
  PanopticFCN & $\mathcal{M}$ & 100\% & 240s & {\bf 80.2} \\
  \bottomrule
\end{tabular}
 \label{tab:abla_anno_detail}
\end{table}

\begin{table*}[t!]

 \caption{Comparisons with previous method on the COCO {\em val} set. Panoptic FCN-400, 512, and 600 denotes utilizing smaller input instead of the default setting. All our results are achieved on the same device with single input and no flipping. FPS is measured {\em end-to-end} from single input to panoptic result with an average speed over 1000 images. The enhanced version is marked with *. The model testing by ourselves according to released codes is denoted as \dag. The window size of SwinT-based model is set to 7.}
 \centering
\begin{tabular}{lcccccccccccc}
  \toprule
  Method & Backbone & PQ & SQ & RQ & PQ$^\mathrm{th}$ & SQ$^\mathrm{th}$ & RQ$^\mathrm{th}$ & PQ$^\mathrm{st}$ & SQ$^\mathrm{st}$ & RQ$^\mathrm{st}$ & Device & FPS\\
  \midrule
  \multicolumn{13}{c}{\small \em box-based} \\
  \midrule
  Panoptic FPN$^\dag$-$\mathrm{1\times}$ & Res50-FPN & 39.4 & 77.8 & 48.3 & 45.9 & 80.9 & 55.3 & 29.6 & 73.3 & 37.7 & V100 & 17.5 \\
  Panoptic FPN$^\dag$-$\mathrm{3\times}$ & Res50-FPN & 41.5 & 79.1 & 50.5 & 48.3 & 82.2 & 57.9 & 31.2 & 74.4 & 39.5 & V100 & 17.5\\
  AUNet~\cite{li2019attention} & Res50-FPN & 39.6 & - & - & 49.1 & - & - & 25.2 & - & - & - & - \\
  CIAE~\cite{gao2020learning} & Res50-FPN & 40.2 & - & - & 45.3 & - & - & 32.3 & - & - & 2080Ti & 12.5 \\
  UPSNet$^\dag$~\cite{xiong2019upsnet} & Res50-FPN & 42.5 & 78.0 & 52.5 & 48.6 & 79.4 & 59.6 & 33.4 & 75.9 & 41.7 & V100 & 9.1 \\
  Unifying~\cite{li2020unifying} & Res50-FPN & 43.4 & 79.6 & 53.0 & 48.6 & - & - & 35.5 & - & - & - & - \\
  \midrule
  \multicolumn{13}{c}{\small \em box-free} \\
  \midrule
  DeeperLab~\cite{yang2019deeperlab} & Xception-71 & 33.8 & - & - & - & - & - & - & - & - & V100 & 10.6 \\
  Panoptic-DeepLab~\cite{cheng2020panoptic} & Res50 & 35.1 & - & - & - & - & - & - & - & - & V100 & 20.0 \\
  AdaptIS~\cite{sofiiuk2019adaptis} & Res50 & 35.9 & - & - & 40.3 & - & - & 29.3 & - & - & - & - \\
  RealTimePan~\cite{hou2020real} & Res50-FPN & 37.1 & - & - & 41.0 & - & - & 31.3 & - & - & V100 & 15.9 \\
  PCV~\cite{wang2020pixel} & Res50-FPN & 37.5 & 77.7 & 47.2 & 40.0 & 78.4 & 50.0 & 33.7 & 76.5 & 42.9 & 1080Ti & 5.7 \\
  SOLO V2~\cite{wang2020solov2} & Res50-FPN & 42.1 & - & - & 49.6  & - & - & 30.7 & - & -  & - & - \\
  Axial-DeepLab~\cite{wang2020axial} & Axial-ResNet-L & 43.6 & - & - & 48.9 & - & - & 35.6 & - & - \\
  MaX-DeepLab~\cite{wang2021max} & MaX-DeepLab-L & 51.1 & - & - & 57.0  & - & - & 42.2 & - & - \\
  \midrule
  Panoptic FCN-400 & Res50-FPN & 40.7 & 80.5 & 49.3 & 44.9 & 82.0 & 54.0 & 34.3 & 78.1 & 42.1 & V100 & {\bf 20.9} \\
  Panoptic FCN-512 & Res50-FPN & 42.3 & 80.9 & 51.2 & 47.4 & 82.1 & 56.9 & 34.7 & 79.1 & 42.7 & V100 & 18.9 \\
  Panoptic FCN-600 & Res50-FPN & 42.8 &	80.6 & 51.6 & 47.9 & 82.6 & 57.2 & 35.1 & 77.4 & 43.1 & V100 & 16.8 \\
  Panoptic FCN & Res50-FPN & 43.6 & 80.6 & 52.6 & 49.3 & 82.6 & 58.9 & 35.0 & 77.6 & 42.9 & V100 & 12.5 \\
  Panoptic FCN$^*$ & Res50-FPN & 44.3 & 80.7 & 53.0 & 50.0 & 83.4 & 59.3 & 35.6 & 76.7 & 43.5 & V100 & 9.2 \\
  \midrule
  Panoptic FCN & SwinT-T-FPN & 47.5 & 81.3 & 56.9 & 53.2 & 83.1 & 63.2 & 39.1 & 78.6 & 47.6 & V100 & 10.8 \\
  Panoptic FCN & SwinT-S-FPN & 49.4 & 82.4 & 58.9 & 55.4 & 83.7 & 65.5 & 40.3 & 80.4 & 48.8 & V100 & 8.7 \\
  Panoptic FCN & SwinT-L-FPN & 51.8 & 83.1 & 61.5 & {\bf 58.6} & 84.4 & {\bf 68.9} & 41.6 & 81.1 & 50.3 & V100 & 4.7 \\
  Panoptic FCN$^*$ & SwinT-L-FPN & {\bf 52.1} & {\bf 83.2} & {\bf 61.6} & 58.5 & {\bf 84.6} & 68.6 & {\bf 42.3} & {\bf 81.1} & {\bf 51.1} & V100 & 4.1 \\
  \bottomrule
\end{tabular}
 \label{tab:coco_val}

\end{table*}

\subsection{Quantitive Results}
We conduct experiments on different scenarios, namely COCO and VOC 2012 for common context, Cityscapes and Mapillary Vistas for traffic-related environments.

\begin{table*}[t]
 \caption{Weakly-supervised results on the COCO and VOC 2012 {\em val} set. $\mathcal{M}$, $\mathcal{B}$, $\mathcal{I}$, and $\mathcal{P}$ indicate mask-, box-, image-, and point-level annotation. Here, VOC 2012 {\em with} COCO represents training and validation on VOC 2012 dataset with COCO pretrained model. All our results are achieved with {\em random} point annotation, single scale input and no flipping.}
\centering
\begin{tabular}{lcccccccccccccc}
  \toprule
  \multirow{2}{*}{Method} & \multirow{2}{*}{Backbone}& \multirow{2}{*}{Supervision} & & \multicolumn{3}{c}{COCO} &  & \multicolumn{3}{c}{VOC 2012} & & \multicolumn{3}{c}{VOC 2012 {\em with} COCO} \\ \cline{5-7} \cline{9-11} \cline{13-15}
  & & & & PQ & PQ$^\mathrm{th}$ & PQ$^\mathrm{st}$ & & PQ & PQ$^\mathrm{th}$ & PQ$^\mathrm{st}$ & & PQ & PQ$^\mathrm{th}$ & PQ$^\mathrm{st}$ \\
  \midrule
  \multicolumn{15}{c}{\small \em fully-supervised} \\
  \midrule
  Li~{\em et.al.}~\cite{li2018weakly} & Res101 & $\mathcal{M}$ & & - & - & - & & 62.7 & - & - & & 63.1 & - & - \\
  DeeperLab~\cite{yang2019deeperlab} & Xception-71 & $\mathcal{M}$ & & 33.8 & - & - & & 67.4 & - & - & & - & - & - \\
  Panoptic-DeepLab~\cite{cheng2020panoptic} & Res50 & $\mathcal{M}$ & & 35.1 & - & - & & - & - & - & & - & - & - \\
  Panoptic FPN~\cite{kirillov2019panopticfpn} & Res50-FPN & $\mathcal{M}$ & & 41.5 & 48.3 & 31.2 & & 65.7 & 64.5 & 90.8 & & - & - & - \\
  Panoptic FCN & Res50-FPN & $\mathcal{M}$ & & {\bf 43.6} & {\bf 49.3} & {\bf 35.0} & & {\bf 67.9} & {\bf 66.6} & {\bf 92.9} & & {\bf 73.1} & {\bf 72.1} & {\bf 93.8} \\
  \midrule
  \multicolumn{15}{c}{\small \em weakly-supervised} \\
  \midrule
  Li~{\em et.al.}~\cite{li2018weakly} & Res101 & $\mathcal{B} + \mathcal{I}$ & & - & - & - & & {\bf 59.0} & - & - & & 59.5 & - & - \\
  JTSM~\cite{shen2021toward} & ResNet18-WS & $\mathcal{I}$ & & 5.3 & 8.4 & 0.7 & & 39.0 & 37.1 & 77.7 & & - & - & - \\
  \midrule
  Panoptic FCN & Res50-FPN & $\mathcal{P}_{10}$ & & 31.2 & 35.7 & 24.3 & & 48.0 & 46.2 & 85.2 & & 52.4 & 50.8 & 86.0 \\
  Panoptic FCN & Res50-FPN & $\mathcal{P}_{20}$ & & 35.1 & 40.0 & 27.7 & & 54.2 & 52.6 & 86.8 & & 58.6 & 57.2 & 87.9 \\
  Panoptic FCN & Res50-FPN & $\mathcal{P}_{30}$ & & {\bf 36.6} & {\bf 41.5} & {\bf 29.2} & & 55.5 & {\bf 53.9} & {\bf 87.8} & & {\bf 60.3} & {\bf 58.9} & {\bf 88.6} \\
  \bottomrule
\end{tabular}
 \label{tab:coco_val_point}

\end{table*}

\subsubsection{COCO}
\noindent
\textbf{Fully-supervised.}
In Table~\ref{tab:coco_val}, we conduct experiments on COCO {\em val} set. 
Compared with recent approaches, Panoptic FCN achieves superior performance with efficiency, which surpasses leading {\em box-based}~\cite{li2020unifying} and {\em box-free}~\cite{wang2020solov2} methods over 0.2\% and {\bf 1.5}\% PQ, respectively. 
With simple enhancement, the gap enlarges to {\bf 0.9}\% and {\bf 2.2}\% PQ. 
Moreover, Panoptic FCN attains a balanced speed-accuracy trade-off, as depicted in Fig.~\ref{fig:speed_acc}.
To compare with recent attention- or transformer-based methods~\cite{wang2020axial,wang2021max}, we further reported the performance that equipped with Swin-Transformer~\cite{liu2021Swin}. As presented in Table~\ref{tab:coco_val}, the network performance improves steadily with stronger SwinT-based models and attains {\bf 52.1}\% PQ with enhanced SwinT-Large model that pretrained on ImageNet-22K dataset~\cite{deng2009imagenet}. 
Meanwhile, Panoptic FCN outperforms all top-ranking models on COCO {\em test-dev} set, as illustrated in Table~\ref{tab:coco_test}. In particular, the proposed method surpasses the state-of-the-art approach in {\em box-based} stream with 0.2\% PQ and attains {\bf 47.5}\% PQ with single scale inputs. Compared with the similar {\em box-free} fashion, our method improves {\bf 1.9}\% PQ over Axial-DeepLab~\cite{wang2020axial} which adopts stronger backbone. As for transformer-based method, Panoptic FCN exceeds recently proposed MaX-DeepLab~\cite{wang2021max} 1.4\% PQ and set a new leading benchmark {\bf 52.7}\% PQ on COCO {\em test-dev} set.

\vspace{0.5em}
\noindent
\textbf{Pointly-supervised.}
To further prove the effectiveness on weakly-supervised setting, we compare with previous method in Table~\ref{tab:coco_val_point}. 
In particular, our point-based method surpasses JTSM~\cite{shen2021toward}, which adopts image-level annotation, with a quite large margin. 
Considering the performance and actual annotation cost, the proposed point-based method is more suitable for practical usage. 
Moreover, the pointly-supervised Panoptic FCN with $\mathcal{P}_{20}$ even achieves comparable performance with fully-supervised Panoptic-DeepLab~\cite{cheng2020panoptic}. 
If the annotation is increased to $\mathcal{P}_{30}$, the performance gap compared with~\cite{cheng2020panoptic} is enlarged to {\bf 1.5}\% PQ.
This proves the effectiveness and efficiency of the pointly-supervised approach, which even exceeds other fully-supervised methods with much less annotation cost.

\begin{table}[th!]
 \caption{Experiments on the COCO {\em test-dev} set. Our results are achieved with single scale and no flipping. The enhanced version and {\em val} set for training are marked with * and \ddag. The window size of SwinT-based model is set to 7.}
 \centering

\begin{tabular}{lcccc}
  \toprule
  Method & Backbone & PQ & PQ$^\mathrm{th}$ & PQ$^\mathrm{st}$\\
  \midrule
  \multicolumn{5}{c}{\small \em box-based} \\
  \midrule
  Panoptic FPN~\cite{kirillov2019panopticfpn}  & Res101-FPN & 40.9 & 48.3 & 29.7 \\
  CIAE~\cite{gao2020learning} & DCN101-FPN & 44.5 & 49.7 & 36.8 \\
  AUNet~\cite{li2019attention} & ResNeXt152-FPN & 46.5 & 55.8 & 32.5 \\
  UPSNet~\cite{xiong2019upsnet} & DCN101-FPN & 46.6 & 53.2 & 36.7 \\
  Unifying$^{\ddag}$~\cite{li2020unifying} & DCN101-FPN & 47.2 & 53.5 & 37.7 \\
  BANet~\cite{chen2020banet} & DCN101-FPN & 47.3 & 54.9 & 35.9 \\
  \midrule
  \multicolumn{5}{c}{\small \em box-free} \\
  \midrule
  DeeperLab~\cite{yang2019deeperlab} & Xception-71 & 34.3 & 37.5 & 29.6 \\
  SSAP~\cite{gao2019ssap} & Res101-FPN & 36.9 & 40.1 & 32.0 \\
  PCV~\cite{wang2020pixel} & Res50-FPN & 37.7 & 40.7 & 33.1 \\
  Panoptic-DeepLab~\cite{cheng2020panoptic} & Xception-71 & 39.7 & 43.9 & 33.2 \\
  AdaptIS~\cite{sofiiuk2019adaptis} & ResNeXt-101 & 42.8 & 53.2 & 36.7 \\
  Axial-DeepLab~\cite{wang2020axial} & Axial-ResNet-L & 43.6 & 48.9 & 35.6 \\
  MaX-DeepLab~\cite{wang2021max} & MaX-DeepLab-L & 51.3 & 57.2 & 42.4 \\
  \midrule
  Panoptic FCN & Res101-FPN & 45.5 & 51.4 & 36.4 \\
  Panoptic FCN & DCN101-FPN & 47.0 & 53.0 & 37.8 \\
  Panoptic FCN$^*$ & DCN101-FPN & 47.1 & 53.2 & 37.8 \\
  Panoptic FCN$^{*\ddag}$ & DCN101-FPN & 47.5 & 53.7 & 38.2 \\
  \midrule
  Panoptic FCN & SwinT-L-FPN & 52.1 & 58.7 & 42.2 \\
  Panoptic FCN$^*$ & SwinT-L-FPN & 52.4 & 59.0 & 42.4 \\
  Panoptic FCN$^{*\ddag}$ & SwinT-L-FPN & {\bf 52.7} & {\bf 59.4} & {\bf 42.5} \\
  \bottomrule
\end{tabular}

 \label{tab:coco_test}
\end{table}
\begin{table}[t]
 \caption{Experiments on the Cityscapes {\em val} set. All our results are achieved with single scale and no flipping. The simple enhanced version is marked with *. The window size of SwinT-based model is set to 7.}
 \centering

\begin{tabular}{lcccc}
  \toprule
  Method & Backbone & PQ & PQ$^\mathrm{th}$ & PQ$^\mathrm{st}$\\
  \midrule
  \multicolumn{5}{c}{\small \em box-based} \\
  \midrule
  Panoptic FPN~\cite{kirillov2019panopticfpn} & Res101-FPN & 58.1 & 52.0 & 62.5 \\
  AUNet~\cite{li2019attention} & Res101-FPN & 59.0 & 54.8 & 62.1 \\
  UPSNet~\cite{xiong2019upsnet} & Res50-FPN & 59.3 & 54.6 & 62.7 \\
  SOGNet~\cite{yang2019sognet} & Res50-FPN & 60.0 & 56.7 & 62.5 \\
  Seamless~\cite{porzi2019seamless} & Res50-FPN & 60.2 & 55.6 & 63.6 \\
  Unifying~\cite{li2020unifying} & Res50-FPN & 61.4 & 54.7 & 66.3 \\
  \midrule
  \multicolumn{5}{c}{\small \em box-free} \\
  \midrule
  PCV~\cite{wang2020pixel} & Res50-FPN & 54.2 & 47.8 & 58.9 \\
  BBFNet~\cite{bonde2020towards} & Res50-FPN & 56.3 & 49.7 & 61.0 \\
  DeeperLab~\cite{yang2019deeperlab} & Xception-71 & 56.5 & - & - \\
  SSAP~\cite{gao2019ssap} & Res50-FPN & 58.4 & 50.6 & - \\
  AdaptIS~\cite{sofiiuk2019adaptis} & Res50 & 59.0 & 55.8 & 61.3 \\
  Panoptic-DeepLab~\cite{cheng2020panoptic} & Res50 & 59.7 & - & - \\
  Axial-DeepLab~\cite{wang2020axial} & Axial-ResNet-L & 63.9 & - & - \\
  Axial-DeepLab~\cite{wang2020axial} & Axial-ResNet-XL & 64.4 & - & - \\
  \midrule
  Panoptic FCN & Res50-FPN & 59.4 & 51.4 & 65.1 \\
  Panoptic FCN$^*$ & Res50-FPN & 61.4 & 54.8 & 66.6 \\
  \midrule
  Panoptic FCN & SwinT-T-FPN & 62.8 & 53.9 & 69.2 \\
  Panoptic FCN & SwinT-S-FPN & 63.3 & 54.2 & 70.0 \\
  Panoptic FCN & SwinT-L-FPN & 64.1 & 55.6 & 70.2 \\
  Panoptic FCN$^*$ & SwinT-L-FPN & {\bf 65.9} & {\bf 59.5} & {\bf 70.6} \\
  \bottomrule
\end{tabular}

 \label{tab:city_val}
\end{table}

\subsubsection{VOC 2012}
\noindent
\textbf{Fully-supervised.} 
In Table~\ref{tab:coco_val_point}, we report results on VOC 2012 dataset. 
As shown in the table, Panoptic FCN achieves much better performance than previous {\em box-based}~\cite{li2018weakly,kirillov2019panopticfpn} and {\em box-free}~\cite{yang2019deeperlab,cheng2020panoptic} methods, and obtains 67.9\% PQ.
If pretrained with COCO dataset, the performance is improved with 5.2\% PQ gain and attains {\bf 73.1}\% PQ based on Res50-FPN. 
Compared with previous method, it surpasses~\cite{li2018weakly} that uses stronger backbone with {\bf 10}\% PQ on VOC 2012 {\em val} set.

\vspace{0.5em}
\noindent
\textbf{Pointly-supervised.}
Similar to that on COCO dataset, we also report the comparison with previous weakly-supervised approaches on VOC 2012 dataset in Table~\ref{tab:coco_val_point}. Compared with the image-supervised method~\cite{shen2021toward}, our proposed manner also achieves much better performance. With COCO dataset for pretraining, the pointly-supervised Panoptic FCN surpasses~\cite{li2018weakly} with 0.8\% PQ that utilizes box- and image-level annotations for supervision.

\subsubsection{Cityscapes} 
Furthermore, we carry out experiments on Cityscapes {\em val} set in Table~\ref{tab:city_val}. As presented in the table, compared with CNN-based networks, Panoptic FCN exceeds the top {\em box-free} model~\cite{cheng2020panoptic} with {\bf 1.7}\% PQ. Even compared with the leading {\em box-based} models~\cite{li2020unifying,chen2020banet}, the proposed method still achieves comparable result. If equipped with SwinT-based backbone, the proposed method achieves much better performance. Specifically, compared with similar attention-based Axial-DeepLab~\cite{wang2020axial}, our proposed method exceeds Axial-ResNet-XL 1.5\% PQ and attains {\bf 65.9}\% PQ. An interesting finding is that the enhanced encoder achieves greater improvement (1.8\% {\em vs.} 0.3\%) than that in COCO.

\subsubsection{Mapillary Vistas} 
In Table~\ref{tab:mapillary_val}, we compare with other state-of-the-art models on the large-scale Mapillary Vistas {\em val} set with CNN- and Transformer-based backbones. As presented in the table, the proposed approach exceeds leading {\em box-free} methods by a large margin in both things and stuff. Specifically, Panoptic FCN surpasses the leading {\em box-based}~\cite{porzi2019seamless} and {\em box-free}~\cite{cheng2020panoptic} models
with 0.7\% and {\bf 3.6}\% PQ with Res50 backbone, and attains {\bf 36.9}\% PQ with simple enhancement in the feature encoder. Given SwinT-based backbone, the proposed network obtains consistent improvement and achieves 43.6\% PQ with SwinT-L. Similar to that in Cityscapes dataset, the simple enhanced encoder brings extra 2.1\% PQ gain, and attains {\bf 45.7}\% PQ on Mapillary Vistas {\em val} set. Compared with attention-based methods, Panoptic FCN improves {\bf 5.6}\% PQ over Axial-DeepLab~\cite{wang2020axial} and performs much better especially in things with {\bf 8.1}\% PQ improvement. It should be noted that only SwinT-based models with window size 7 are used in the experiments, and a larger window size like 12 could bring much better results.

\subsection{Qualitative Results}
We further visualize results of Panoptic FCN on several datasets with {\em common context} and {\em traffic-related} scenarios.

\begin{table}[t!]
 \caption{Experiments on the Mapillary Vistas {\em val} set. All our results are achieved with single scale and no flipping. The enhanced version is marked with *. The window size of SwinT-based model is set to 7.}
\centering
\begin{tabular}{lcccc}
  \toprule
  Method & Backbone & PQ & PQ$^\mathrm{th}$ & PQ$^\mathrm{st}$\\
  \midrule
  \multicolumn{5}{c}{\small \em box-based} \\
  \midrule
  BGRNet~\cite{wu2020bidirectional} & Res50-FPN & 31.8 & 34.1 & 27.3 \\
  TASCNet~\cite{li2018learning} & Res50-FPN & 32.6 & 31.1 & 34.4 \\
  Seamless~\cite{porzi2019seamless} & Res50-FPN & 36.2 & 33.6 & 40.0 \\
  \midrule
  \multicolumn{5}{c}{\small \em box-free} \\
  \midrule
  DeeperLab~\cite{yang2019deeperlab} & Xception-71 & 32.0 & - & - \\
  AdaptIS~\cite{sofiiuk2019adaptis} & Res50 & 32.0 & 26.6 & 39.1 \\
  Panoptic-DeepLab~\cite{cheng2020panoptic} & Res50 & 33.3 & - & - \\
  Panoptic-DeepLab~\cite{cheng2020panoptic} & Xception-71 & 37.7 & - & - \\
  Axial-DeepLab~\cite{wang2020axial} & Axial-ResNet-L & 40.1 & 32.7 & 49.8 \\
  \midrule
  Panoptic FCN & Res50-FPN & 34.8 & 30.6 & 40.5 \\
  Panoptic FCN$^*$ & Res50-FPN & 36.9 & 32.9 & 42.3 \\
  \midrule
  Panoptic FCN & SwinT-T-FPN & 39.0 & 33.4 & 46.4 \\
  Panoptic FCN & SwinT-S-FPN & 41.2 & 35.4 & 48.8 \\
  Panoptic FCN & SwinT-L-FPN & 43.6 & 38.3 & 50.6 \\
  Panoptic FCN$^*$ & SwinT-L-FPN & {\bf 45.7} & {\bf 40.8} & {\bf 52.1} \\
  \bottomrule
\end{tabular}
 \label{tab:mapillary_val}

\end{table}

\begin{figure*}[tp!] 

\centering
\subfigure[Visualization on COCO {\em val} set.]{
\includegraphics[width=0.9\linewidth]{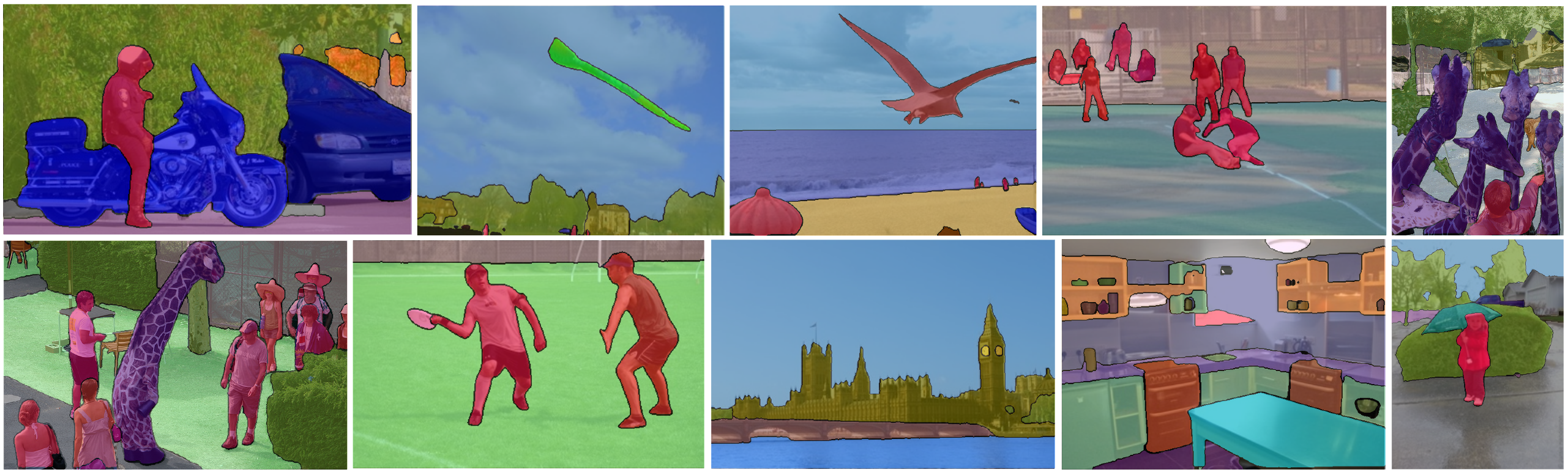}
\label{fig:vis_coco}
}
\subfigure[Visualization on VOC 2012 {\em val} set.]{
\includegraphics[width=0.9\linewidth]{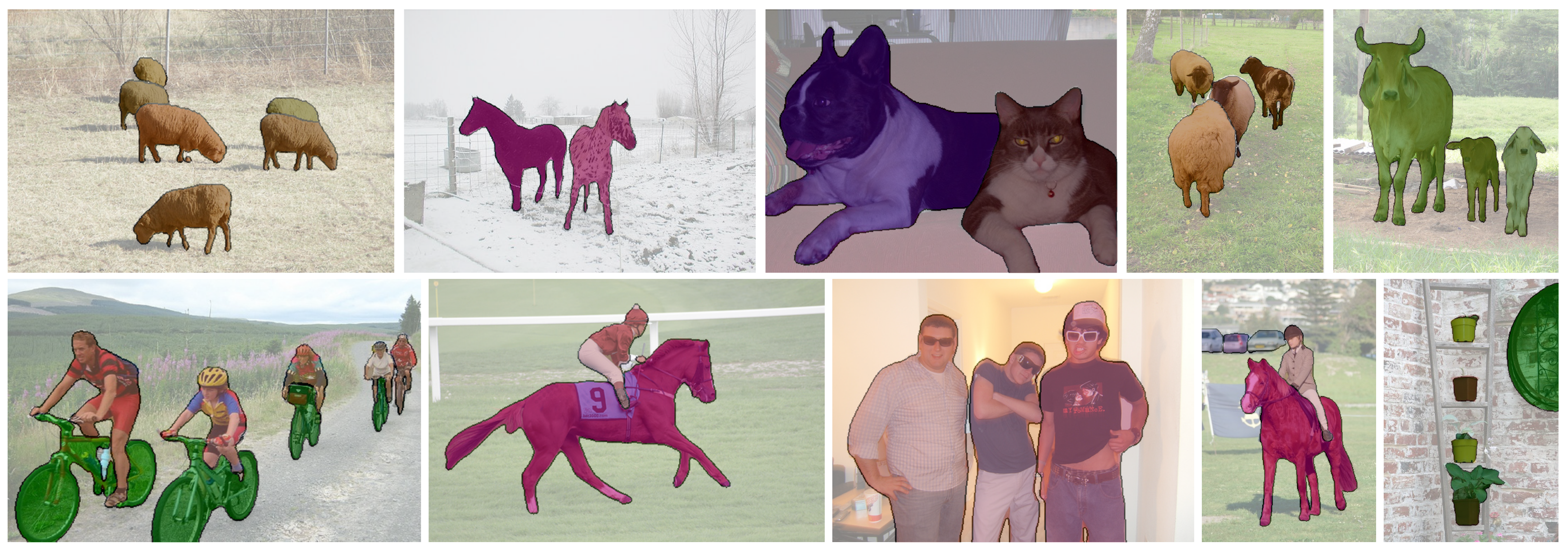}
\label{fig:vis_voc}
}
\subfigure[Visualization on Cityscapes {\em val} set.]{
\includegraphics[width=0.9\linewidth]{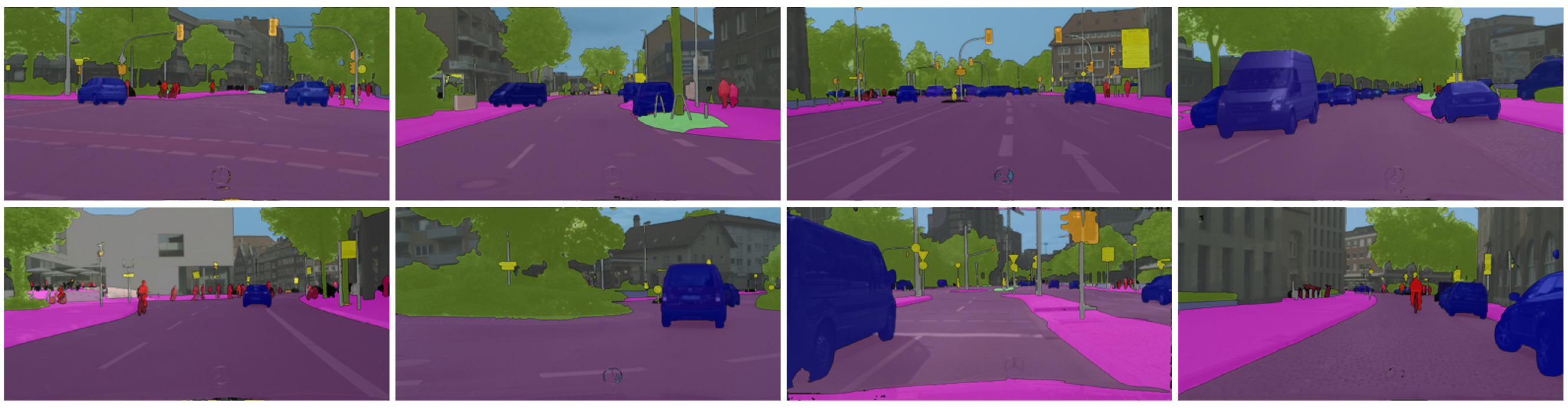}
\label{fig:vis_city}
}
\subfigure[Visualization on Mapillary Visita {\em val} set.]{
\includegraphics[width=0.9\linewidth]{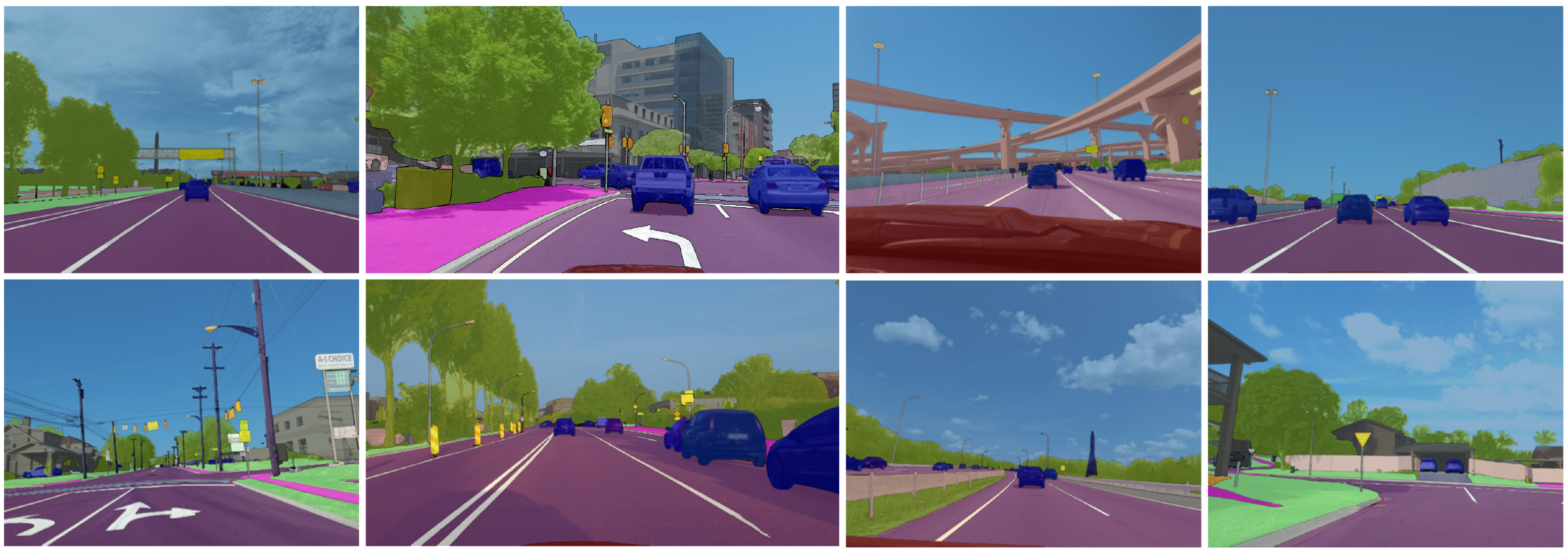}
\label{fig:vis_mapillary}
}
\caption{Qualitative results of Panoptic FCN on four widely-adopted datasets.}
\vspace{1em}
\label{fig:visualize}
\end{figure*}

\vspace{0.5em}
\noindent
{\bf COCO.} As presented in Fig.~\ref{fig:vis_coco}, Panoptic FCN gives detailed characterization to common environment. Thanks to the {\em pixel-by-pixel} manner and unified representation, details in both things and stuff can be preserved. 

\vspace{0.5em}
\noindent
{\bf VOC 2012.} In Fig.~\ref{fig:vis_voc}, we present results on VOC 2012 {\em val} set. The proposed approach validates its effectiveness on objects with various scales, where edges are preserved well.

\vspace{0.5em}
\noindent
{\bf Cityscapes.} As for the street view, we visualize results on the Cityscapes {\em val} set in Fig.~\ref{fig:vis_city}. In addition to the well-depicted cars and pedestrians, the proposed approach achieves satisfactory performance on slender objects.

\vspace{0.5em}
\noindent
{\bf Mapillary Vistas.} In Fig.~\ref{fig:vis_mapillary}, we present panoptic results on the Mapillary Vistas {\em val} set, which contains larger scales scenes. It is clear that the proposed method achieves surprising results, especially on vehicles and traffic signs.

\section{Conclusion}~\label{sec:conclusion}
We have presented the Panoptic FCN, a conceptually simple yet effective framework for fully- and weakly-supervised panoptic segmentation. The key difference from prior works lies in that we represent and predict things and stuff in a fully convolutional manner. To this end, {\em kernel generator} and {\em kernel fusion} are proposed to generate the unique kernel weight for each object instance or semantic category. With the high-resolution feature produced by {\em feature encoder}, prediction is achieved by convolutions directly. Thus, {\em instance-awareness} and {\em semantic-consistency} for things and stuff are respectively satisfied with the designed workflow.
Meanwhile, a unified {\em point-based} manner is proposed to provide supervision for both things and stuff. With the designed annotation and generation strategies, the pointly-supervised version maintains the high performance but save tremendous human annotation cost.

\bibliographystyle{IEEEtran}
\bibliography{IEEEabrv,panopticfcn}

\end{document}